\documentclass[examplefnt,biber]{format/nowfnt} 

\usepackage[utf8]{inputenc}
\usepackage{subcaption}
\usepackage{enumitem}
\usepackage{amsfonts,amsmath,amssymb}
\usepackage[framemethod=tikz]{mdframed}
\usepackage{comment}

\newcommand{\etc}{\emph{etc}}

\title{A Survey on the Integration of Machine Learning with Sampling-based Motion Planning}

\maintitleauthorlist{
Troy McMahon$*$ \\
Rutgers University \\
tm799@rutgers.edu
\and
Aravind Sivaramakrishnan$*$ \\
Rutgers University \\
as2578@rutgers.edu
\and
Edgar Granados \\
Rutgers University \\
eg585@rutgers.edu
\and
Kostas E. Bekris \\
Rutgers University \\
kb572@rutgers.edu
}

\issuesetup
{%
 copyrightowner={T.~McMahon \textit{et al.}},
 volume        = 9,
 issue         = 4,
 pubyear       = 2022,
 isbn          = xxx-x-xxxxx-xxx-x,
 eisbn         = xxx-x-xxxxx-xxx-x,
 doi           = 10.1561/2300000063,
 firstpage     = 266, 
 lastpage      = 327
 }
 
\usepackage{mwe}

\author[1,*]{McMahon, Troy}
\author[1,*]{Sivaramakrishnan, Aravind}
\author[1]{Granados, Edgar}
\author[1]{Bekris, Kostas E.}

\affil[1]{Department of Computer Science, Rutgers University, NJ, USA; kb572@rutgers.edu, as2578@rutgers.edu}
\affil[*]{Equal contribution.}

\articledatabox{\nowfntstandardcitation}

\newenvironment{myitem}{\begin{list}{$\bullet$}
{\setlength{\topsep}{0pt}
\setlength{\labelwidth}{0pt}
\setlength{\leftmargin}{10pt}
\setlength{\parsep}{-0pt}
\setlength{\itemsep}{1pt}
\setlength{\partopsep}{0pt}}}%
{\end{list}}

\newenvironment{myenum}{\begin{list}{$\bullet$}
{\setlength{\topsep}{0pt}
\setlength{\labelwidth}{0pt}
\setlength{\leftmargin}{5pt}
\setlength{\parsep}{-0pt}
\setlength{\itemsep}{1pt}
\setlength{\partopsep}{0pt}}}%
{\end{list}}

\newcommand{\sbmp}{{\tt SBMP}}
\newcommand{\sbmps}{{\tt SBMPs}}
\newcommand{\cspace}{\mathbb{C}}

\newcommand{\edgar}[1]{ {\color{purple}#1} }

\begin{document}

\makeabstracttitle

\begin{abstract}
Sampling-based methods are widely adopted solutions for robot motion planning. The methods are straightforward to implement, effective in practice for many robotic systems. It is often possible to prove that they have desirable properties, such as probabilistic completeness and asymptotic optimality. Nevertheless, they still face challenges as the complexity of the underlying planning problem increases, especially under tight computation time constraints, which impact the quality of returned solutions or given inaccurate models. This has motivated machine learning to improve the computational efficiency and applicability of Sampling-Based Motion Planners (\sbmps). This survey reviews such integrative efforts and aims to provide a classification of the alternative directions that have been explored in the literature. It first discusses how learning has been used to enhance key components of \sbmps, such as node sampling, collision detection, distance or nearest neighbor computation, local planning, and termination conditions. Then, it highlights planners that use learning to adaptively select between different implementations of such primitives in response to the underlying problem's features.  It also covers emerging methods, which build complete machine learning pipelines that reflect the traditional structure of \sbmps. It also discusses how machine learning has been used to provide data-driven models of robots, which can then be used by a \sbmp. Finally, it provides a comparative discussion of the advantages and disadvantages of the approaches covered, and insights on possible future directions of research. An online version of this survey can be found at: \url{https://prx-kinodynamic.github.io/}
\end{abstract}

\chapter{Introduction}
\label{sec:intro}

\begin{figure}[t!]
    \centering
    \begin{subfigure}{\linewidth}
    \centering
    \includegraphics[width=5.0cm,height=5.0cm]{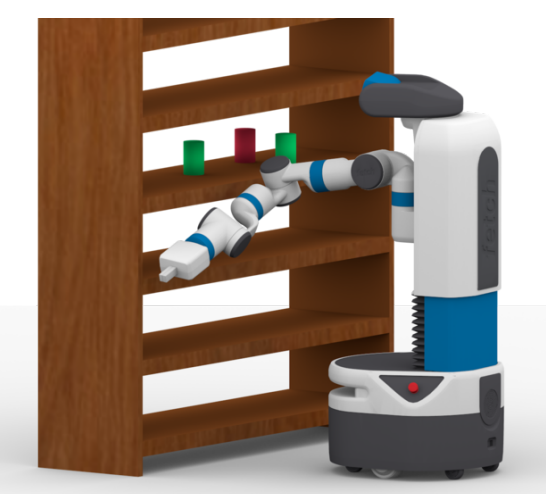}
    \includegraphics[width=5.0cm,height=5.0cm]{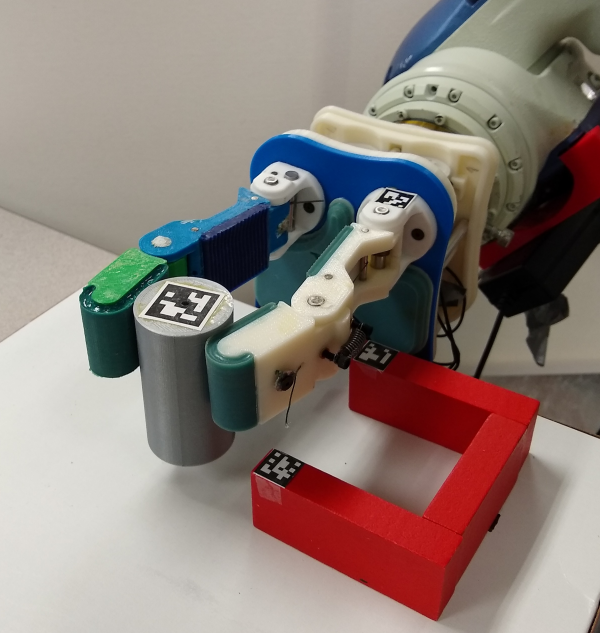}
    \end{subfigure} \\
    \vspace{0.1cm}
    \begin{subfigure}{\linewidth}
    \centering
    \includegraphics[width=5.0cm,height=5.0cm]{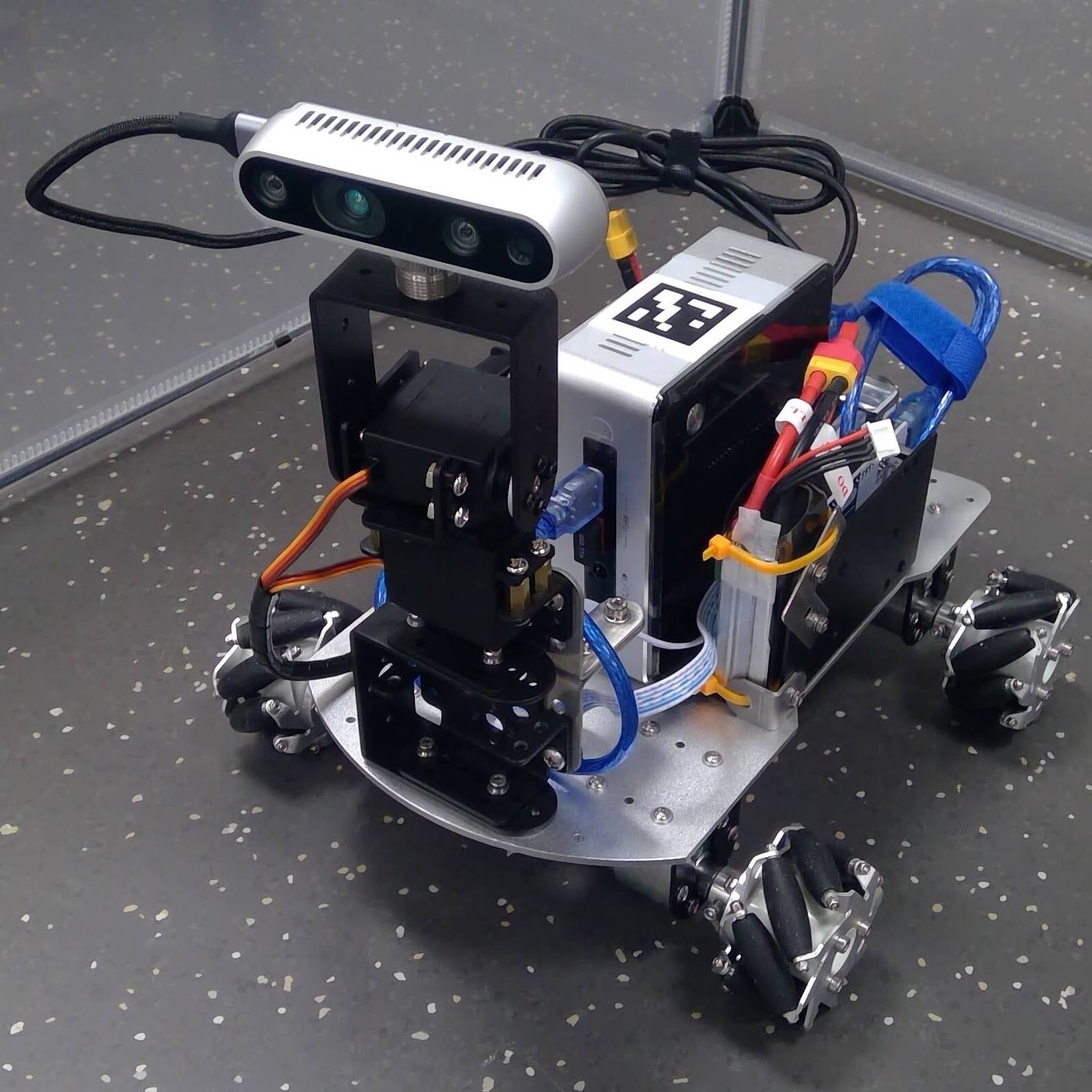}
    \includegraphics[width=5.0cm,height=5.0cm]{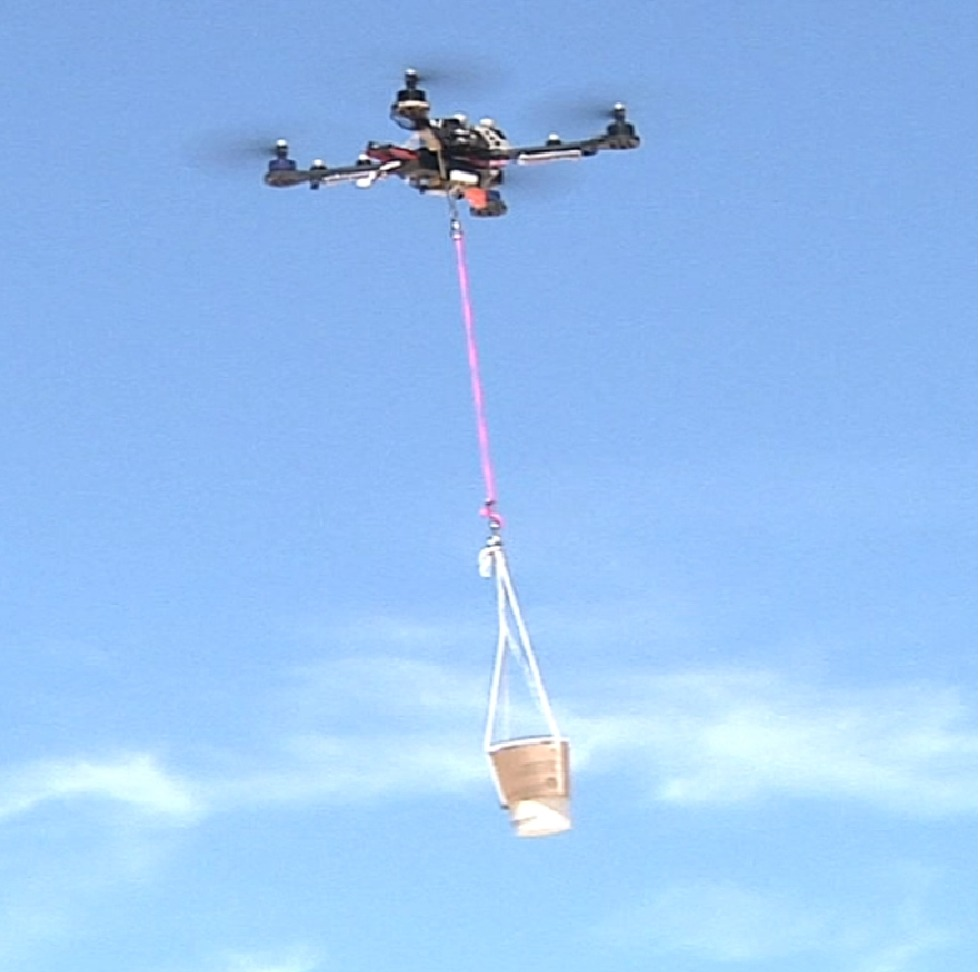}
    \end{subfigure} \\
    \vspace{-.05in}
    \caption{\textbf{Example domains where machine learning has been integrated with sampling-based motion planning:}  a) grasping objects with robotic arms (image from \cite{chamzas2019using}),  b) dexterous manipulation with adaptive hands (image from \cite{9196564}),
     c) robot navigation of ground robots (image from \cite{omnirobot}),
     d) aerial vehicles (image from \cite{FPCFT-2017}).}   
    \vspace{-.25in}
    \label{fig:introduction}
\end{figure}

Motion planning is the problem of finding valid paths, expressed as sequences of configurations, or trajectories, expressed as sequences of controls, which move a robot from a given start state to a desired goal state while avoiding obstacles.  It has applications in problems ranging from mobile robotics \citep{Barraquand_1991}, manipulation planning \citep{OM-2005}, graphics and animation \citep{KAAT-2008}, protein folding \citep{SA-2001} to crowd simulation \citep{bayazit2002better,sud_2008,Toll-2012} and multi-robot applications \citep{svestka-1995,Clark2001RandomizedMP}.  Variations of the motion planning problem can include dynamic constraints, which can be important in autonomous driving and aerial vehicles (Figure~\ref{fig:introduction}).  

The motion planning problem is $\texttt{PSPACE-Complete}$ \citep{reif1979complexity,canny1988complexity,latombe1991robot}, and its complexity depends exponentially on the number of degrees of freedom of the robotic system.  This makes traditional, complete methods difficult to apply for problems with more than 4 or 5 degrees of freedom.  This has motivated work on developing sampling-based methods, which often use random samples to explore the underlying configuration space. Examples of popular Sampling-Based Motion Planners (\sbmps) include the Probabilistic Roadmap ({\tt PRM}) \citep{KSLO-1996}, the Rapidly-exploring Random Tree ({\tt RRT}) \citep{L-1998} and the Expansive Spaces \citep{hsu1997path} algorithms.  Sampling-based motion planners give up on the traditional notion of completeness and instead aim for probabilistic completeness, which means that they are guaranteed to eventually discover a solution if one exists but cannot confirm solution non-existence.  Progress in the field has also allowed the development of methods, such as {\tt RRT$^*$} and {\tt PRM$^*$} \citep{KF-2011}, which are also asymptotically optimal, i.e., they guarantee convergence to an optimal solution if one exists.

Beyond their algorithmic properties, these methods have proven quite effective in finding solutions for relatively high-dimensional challenges, where the traditional approaches do not scale. Their popularity also stems from the fact that they provide flexible frameworks, which are rather straightforward to implement and adapt for a large variety of robotic systems. Nevertheless, they may still face challenges as the complexity of the underlying planning problem increases:

\begin{myenum}
\item[1.] Some of the challenges relate to \emph{computational efficiency}, which may be hindered from the exploration of the underlying configuration space via sampling, especially when the key primitives of these planners, such as collision checking or forward propagation of the system's dynamics, are computationally expensive. 
\item[2.] Other issues relate to \emph{path quality}. Despite the progress in understanding the conditions for asymptotic optimality, convergence to high-quality solutions may be hindered in practice when naive exploration primitives are employed, such as the random sampling of controls. 
\item[3.] Furthermore, \sbmps\, like most motion planning methods, typically assume the \emph{availability of an accurate, complete model}. Traditional, engineered models may be inaccurate or unable to express all critical physical aspects of the problem or not predict how a dynamic environment may evolve. Furthermore, sensing constraints may introduce partial observability and uncertainty about the environment. These factors can limit the applicability of \sbmps. 
\end{myenum}

\noindent These challenges motivate the use of machine learning to improve the computational efficiency of \sbmps, accelerate their practical convergence to high-quality solutions, and provide access to accurate-enough, data-driven models, which adapt to varying environmental conditions and sensing input.

Machine learning enables the autonomous derivation of solutions to problems based on prior experience and data.  It promises to constantly improve performance by incorporating new data and identifying solutions that engineered approaches may not be able to achieve. This makes machine learning especially useful for an application like robotics, where a robotic agent must contend with an endless variety of tasks and environments.  
There are also many scenarios where learned agents can approximate costly computations.  In these cases, the learned agent can be trained to model the computations in a pre-processing step, then used during run-time in place of the computation or as a heuristic for it.  This is especially useful in robotics, where the robot must act and react to situations in real-time.         

Fundamentally, machine learning algorithms operate by building a model of observed data, that can predict and generalize to new examples. This data can come from various sources: an existing dataset, through human training or demonstration, or it can be accumulated from the results of previous predictions that the model has made. For model-based agents, learning is done by fitting the model's parameters to data, which can be done using methods such as regression or reinforcement learning.  During query time, the model is queried to give predictions based on patterns observed in the data.     

There is a large body of literature on the application of machine learning algorithms to improve the efficiency of robotic systems in general \citep{kbp-2013, kroemer2019review}. Recently, there has been a lot of attention on the progress of deep learning methods, which has resulted in many efforts to utilize the corresponding tools in robotics \citep{sunderhauf2018limits}. This survey focuses specifically on integrating machine learning tools to improve the efficiency, convergence, and applicability of \sbmps.

This survey covers a wide breadth of robotic applications, including, but not limited to, mobile robot navigation, manipulation planning, and planning for systems with dynamic constraints. In particular, this monograph first reviews the attempts to use machine learning to improve the performance of individual primitives used by \sbmps (Section~\ref{sec:learning-primitives}). It also studies a series of planners that use machine learning to adaptively select from a set of motion planning primitives. It then proceeds to study a series of integrated architectures that learn an end-to-end mapping of sensor inputs to robot trajectories or controls (Section~\ref{sec:integrated}). Finally, it studies how \sbmps\ can operate over learned models of robotic system that account for noise and uncertainty (Section~\ref{sec:planning_under_uncertainties}).


The survey concludes with a comparative discussion of the different approaches covered in Sections~\ref{sec:learning-primitives}~-~\ref{sec:planning_under_uncertainties}.  It evaluates these approaches in terms of their impact on computational efficiency of the planner, quality of the computed paths, and their overall applicability.  It then outlines the broad difficulties and limitations of these methods, as well as potential directions of future work.

\chapter{Sampling-based Motion Planning}
\label{sec:motion-planning}

\begin{figure*}[h]
    \centering
    \begin{subfigure}{0.352\textwidth}
    \includegraphics[width=0.98\textwidth]{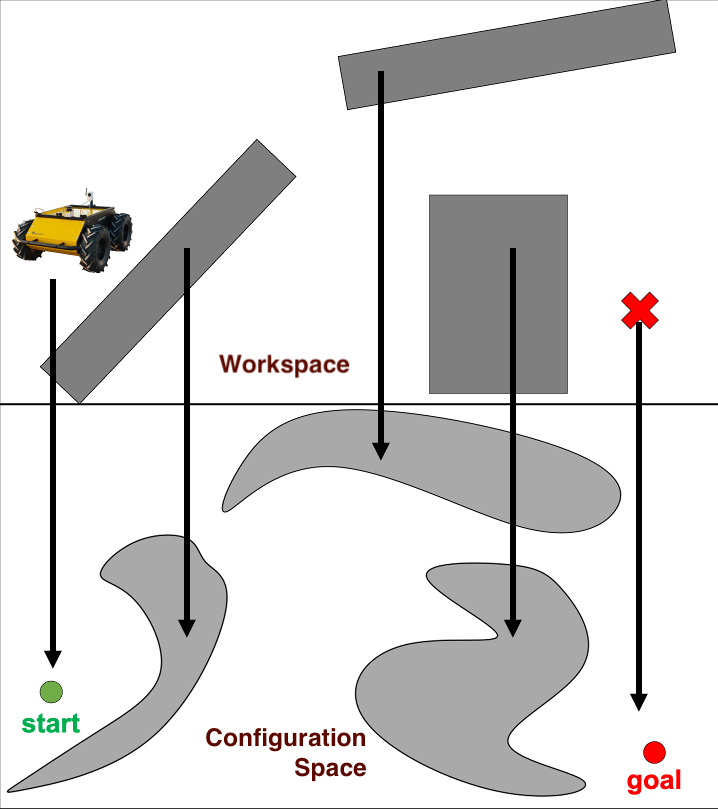}
    \subcaption[]{}
    \end{subfigure}
    \begin{subfigure}{0.638\textwidth}
    \begin{subfigure}{0.49\textwidth}
    \includegraphics[width=\textwidth]{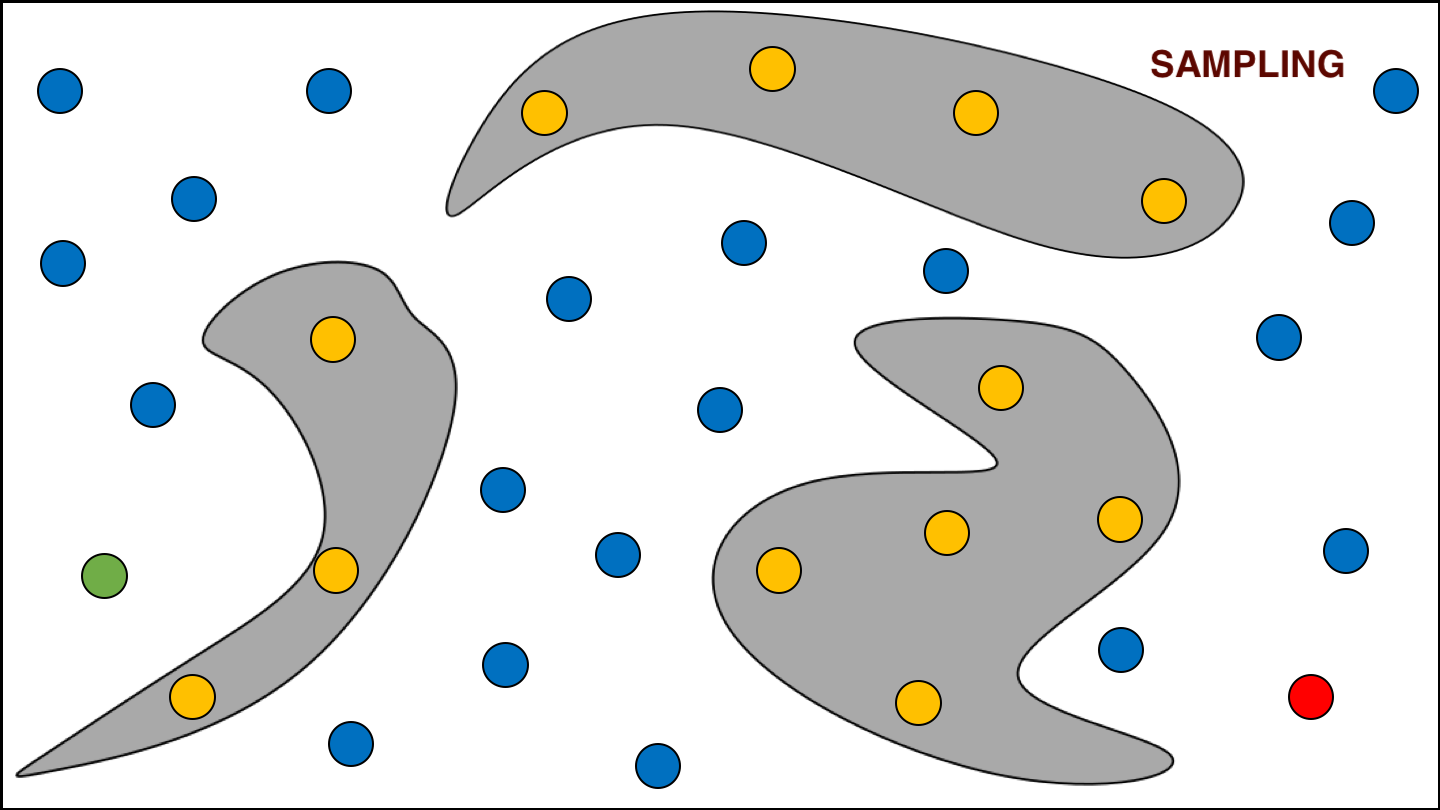}
    \subcaption[]{}
    \end{subfigure}
    \begin{subfigure}{0.49\textwidth}
    \includegraphics[width=\textwidth]{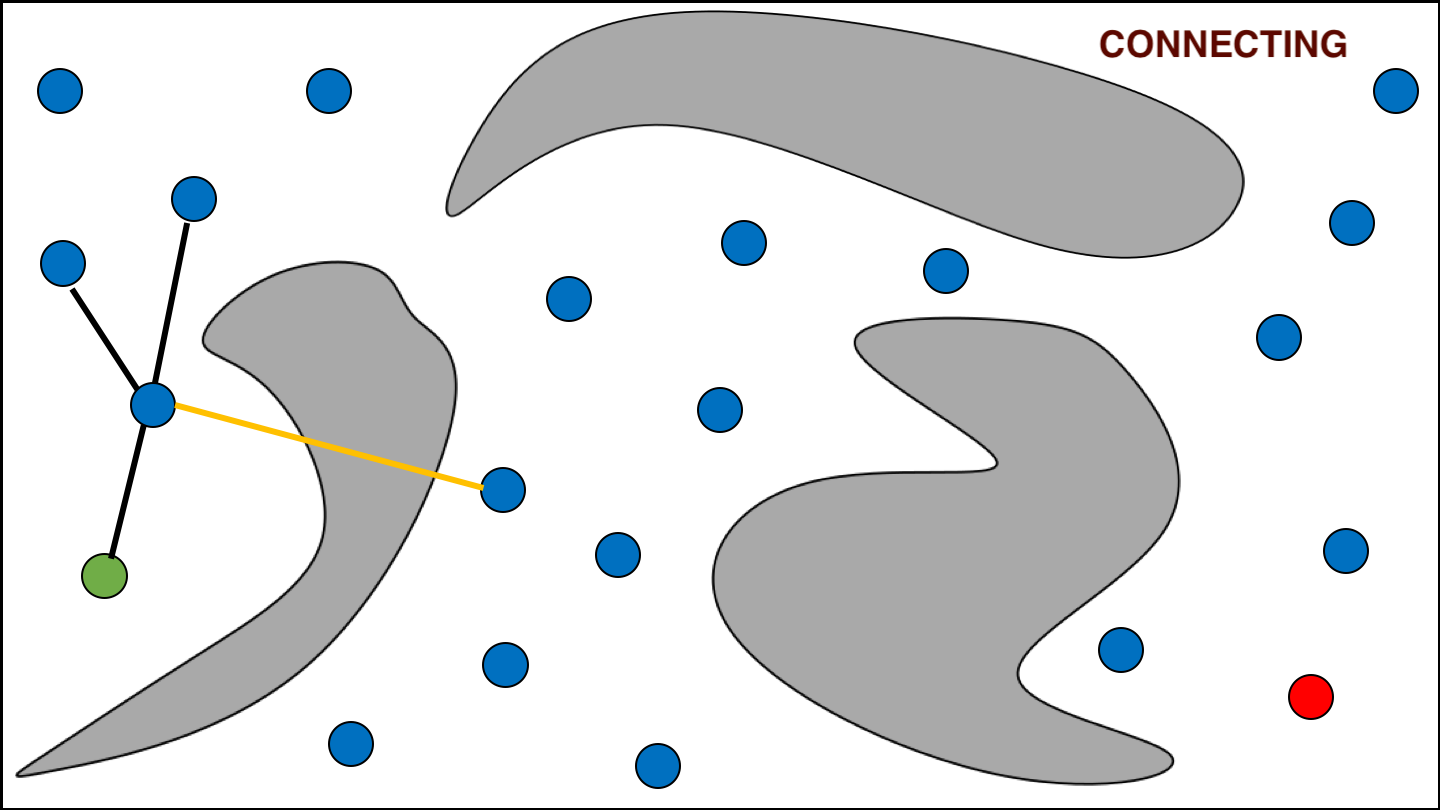}
    \subcaption[]{}
    \end{subfigure}
    \hfill
    \begin{subfigure}{0.49\textwidth}
    \includegraphics[width=\textwidth]{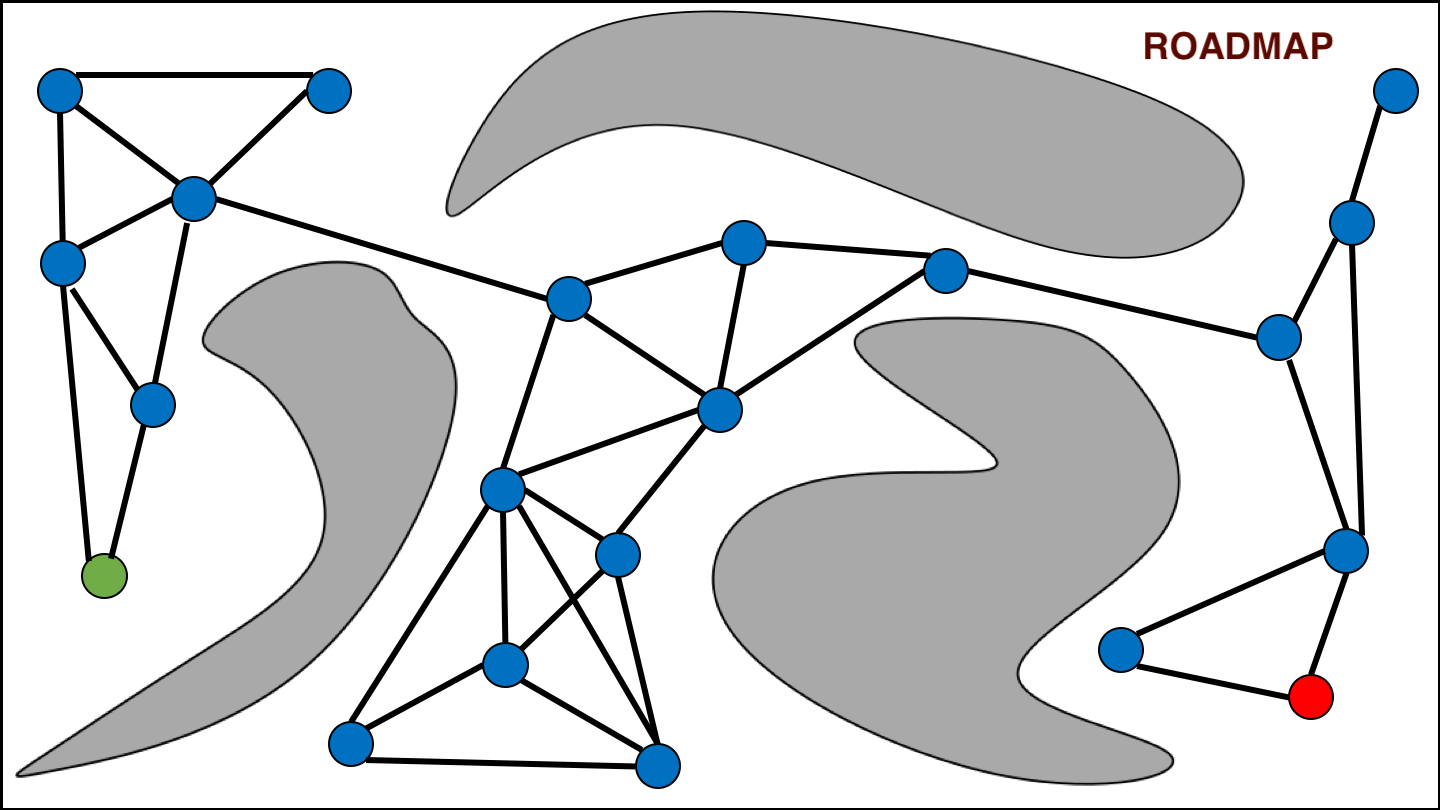}
    \subcaption[]{}
    \end{subfigure}
    \begin{subfigure}{0.49\textwidth}
    \includegraphics[width=\textwidth]{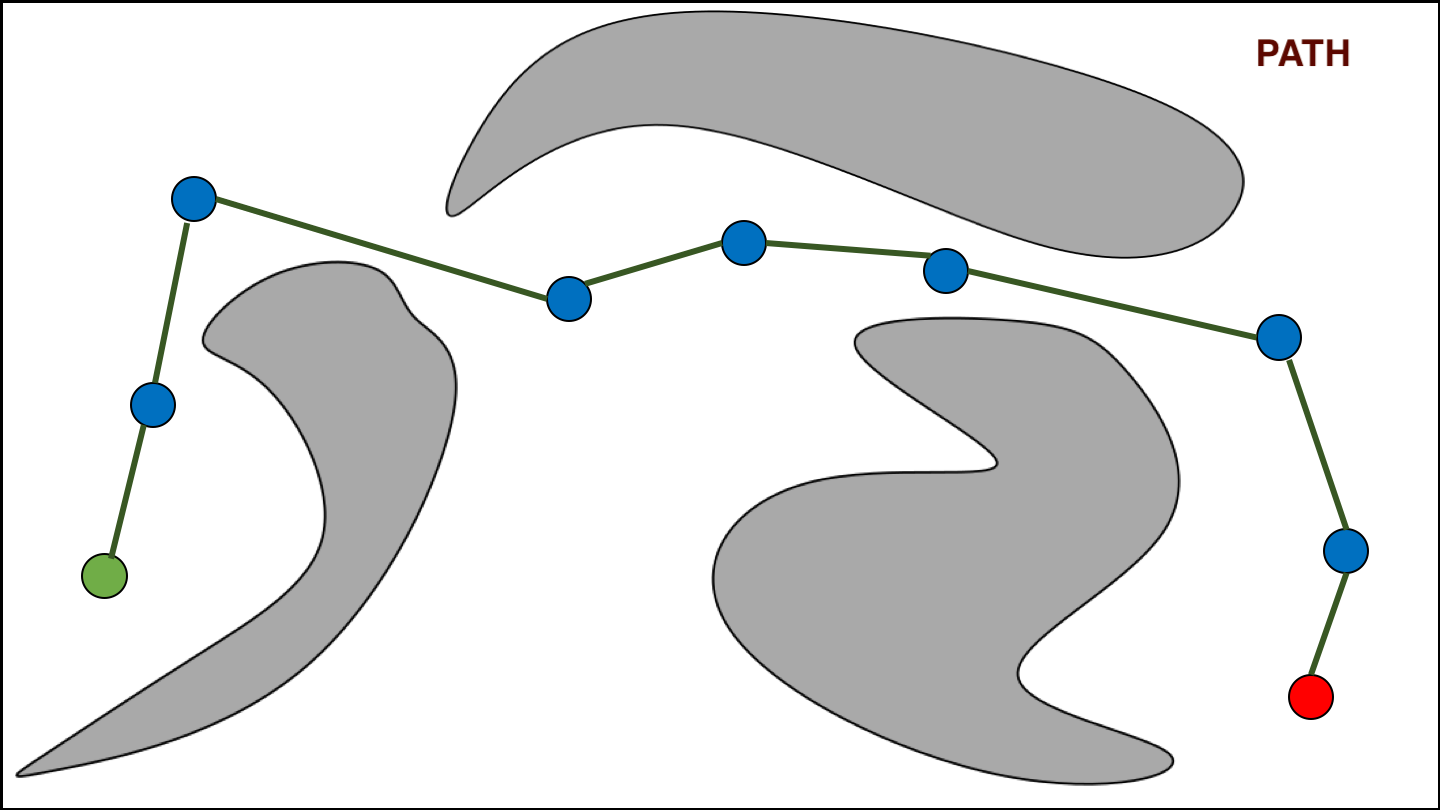}
    \subcaption[]{}
    \end{subfigure}
    \end{subfigure}
    \caption{Solving a motion planning problem using a batch version of the \textit{Probabilistic Road-Map (PRM)} method. (a) The problem is transformed from the \textit{workspace} where the robot has geometry (top) to the \textit{configuration space} (bottom), where the robot is a point. (b) After \textit{sampling} the configuration space and \textit{validity checking} each sample in the workspace, valid samples (blue) are retained and invalid samples (orange) are discarded. (c) For every sample, local edges are \textit{constructed} to nearby samples and \textit{validity checked}. Valid edges (black) are retained and invalid edges (orange) are discarded. (d) Once the final roadmap is obtained, each edge is weighted by a \textit{cost} or a \textit{distance function}. (e) The final path is obtained using a graph search technique (such as A*), typically aided by an admissible \textit{heuristic function}. }
    \label{fig:sbmp-primitives}
\end{figure*}

An instance of a motion planning problem consists of a mobile object in a known environment.  The object could be a geometric shape, a rigid body, or a set of rigid bodies that are connected by joints, such as a robotic arm or articulated linkage.  The environment consists of a geometric space (called the \textit{workspace}), as well as a notion  of \emph{validity} that determines if paths within the environment are valid.  Normally this is accomplished by dividing the environment into valid (i.e. free) regions and invalid regions (i.e. obstacles).  A path is considered to be valid if it is totally contained in the free regions. For problems with constraints (e.g. dynamics constraints), the path must also satisfy these constraints.  A solution to the motion planning problem is a valid path by which the object traverses from a given start configuration to a given goal region.  

The basic version of the motion planning problem assumes that the environment is fully known and that the motions of the robot are exact.  It does not take into account uncertainty in the environment due to issues such as noise in the robot's sensors or actuators.  It also assumes that the robot's location in the environment is always known and as such does not address the problem of robot localization.

The object's motion is defined in terms of degrees of freedom, which include translational and rotational degrees of freedom as well as internal degrees of freedom (e.g. joint angles).  A \emph{configuration} consists of a setting for each of the degrees of freedom of the object and uniquely defines a \textit{state} of the robot.  The space of all possible configurations for a system is defined as the \emph{configuration space} or $\cspace$. A configuration is defined as valid or invalid based on if it satisfies a problem-specified validity condition.  This condition usually requires that there are no collisions between the object and any obstacles in the environment.  For robots that consist of multiple bodies (e.g. articulated linkages), validity conditions usually also require that there is no collisions between the bodies of the robot (i.e. self-collision).  Configurations can be tested for collision using a closed-box collision checker.

A path or a trajectory is a continuous sequence of configurations.  A path/trajectory is valid if all of the configurations along it are valid, otherwise it is invalid.  For systems with constraints (e.g. dynamics), trajectories must also adhere to these constraints of the system in order to be considered valid.  As stated previously, a solution to the motion planning problem is a valid path/trajectory between the start and goal configuration.  

For some applications, it is also important to consider path quality.  The quality of a path is indicated by a given cost function which the problem wishes to minimize (e.g. distance traveled or energy expended by the robot).  A cost function takes as input a path and returns a real number indicating the cost associated with the path.






The complexity of motion planning in $\cspace$ grows exponentially with respect to the number of degrees of freedom, which make modeling $\cspace$ directly computationally challenging.  This has led to the development of sampling-based methods, such as the PRM \citep{KSLO-1996}, RRT \citep{L-1998} and RRT* \citep{KF-2011}, which construct an abstract representation of $\cspace$ via sampling. The sampled configurations are classified as valid (free) or invalid (obstacle) using a validity checker that tests if the sample is in the valid portion of $\cspace$.  Valid samples provide a representation of the free portion of $\cspace$ ($\mathbb{C}_{free}$) and the invalid samples provide a representation of the invalid portion ($\mathbb{C}_{obs}$).  \sbmps \ have been applied to a wide variety of problems including in mobile robot navigation, robotic manipulation, molecular simulations and animation. 

Sampling-based motion planning methods search for paths by constructing a \emph{roadmap} or a \emph{tree} that consists of a set of sampled states (nodes) that are connected by a path/trajectory that is defined by a reproducible function (e.g. straight-line interpolation).  These paths / trajectories are referred to as local plans. Local plans can be tested for validity by stepping through the path/trajectory at a small interval and testing configurations for validity.


The PRM algorithm \citep{KSLO-1996} constructs a roadmap by randomly sampling configurations across $\cspace$.  It then tests the samples for validity using a validity checker (e.g., a collision detector) and removes any samples that are invalid.  It then constructs a roadmap in which there is a node for each valid sample.  It then attempts to connect each sample to its $k$ nearest neighbors according to a distance metric (e.g. Euclidean distance).  If the connection is successful, it adds an edge between the nodes of the roadmap that correspond to the sample and the neighbor.  This roadmap can then be queried by inserting a given start and goal node into the roadmap and using a search procedure like Dijkstra's algorithm to find a path from the start node to the goal node. An illustration of this procedure is provided in Figure~\ref{fig:sbmp-primitives}.

The RRT algorithm \citep{L-1998} builds a tree by incrementally expanding existing nodes.  This method initializes the tree to contain the specified start node.  It then incrementally expands the tree by randomly sampling in $\cspace$, locating the node in the tree that is closest to the sample according to some distance metric and expanding that node by $\epsilon$ in the direction of the sample.  If the local path between the nearest node and the expanded node is valid then the expanded node is added to the tree. The algorithm then tests if the local path terminates in the goal region.  If so, it extracts a path from the start to the goal by backtracking.


Recently there has been a great deal of focus on developing methods that produce good quality paths and in particular on methods that are guaranteed to converge to a good quality path.  This has lead to the advent of  asymptotically optimal motion planners planner RRT*, PRM* \citep{KF-2011}, RRT-\# \citep{arslan2013use}, Stable Sparse RRT \citep{li2014asymptotically}, RRT-Blossom \citep{KP-2006}, Fast Marching Trees (FMT*) \citep{fmt}, AO-X \citep{Hauser_2016},  Dominance-Informed Region Trees (DIRT) \citep{LB-DIRT}, AO-RRT2 \citep{kleinbort2019refined}, and Batch-Informed Trees (BIT*) \citep{GBS-2020}. 
Asymptotically optimal methods are guaranteed to converge to an optimal solution as the number of samples approaches infinity (see related reviews on the subject: \cite{Bekris2020,gammell2021asymptotically}).

RRT* \citep{KF-2011} is an incremental approach that constructs a tree rooted at the start configuration, similar to RRT. At every iteration, the nearest tree node from a sample is used to \textit{steer} towards the sample to generate a node to add. For each such new node, the best parent is selected from a valid \textit{neighborhood} and steering is again used to connect the best parent to the new node. The best parent is decided in terms of the cost from the root of the tree to the new node via the candidate parent. The neighborhood is also \textit{rewired} by checking connections from the new node to existing nodes inside the neighborhood. This ensures that the algorithm maintains the \textit{optimal tree} in terms of cost from the start for the set of generated nodes. The solution can be returned by tracing the tree backwards from a node inside the goal to the root.
\chapter{Learning Primitives of\\ Sampling-based Motion Planning}
\chaptermark{Learning Primitives of SBMP}
\label{sec:learning-primitives}

Sampling-based motion planners, such as PRM \citep{KSLO-1996} and RRT \citep{L-1998}, are comprised of a set of primitive operations such as sample generation, collision detection and distance computation.  This section discusses how machine learning has been applied to each of these primitive operations. It also discusses a set of adaptive methods that use learning to select between different combinations of primitive operations.

\section{Sampling Sequences}
\label{sec:sampling}

\begin{figure}[h!]
    \centering
    \includegraphics[width=\textwidth]{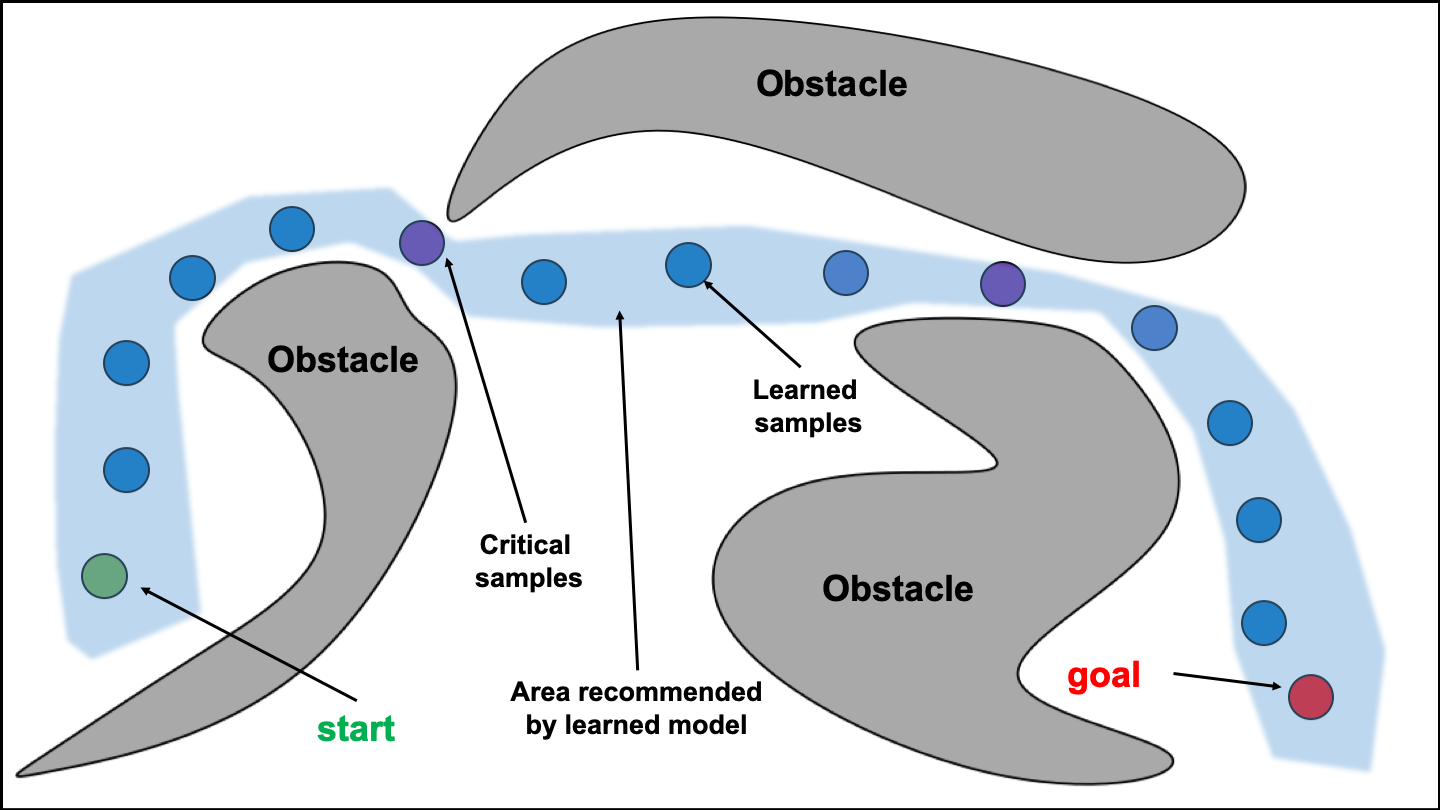}
    \caption{Given a planning problem: start (red), goal (green) and obstacles (grey), a learned sampling strategy predicts samples inside the blue region, characterized by parts of $\mathbb{C}_{free}$ that lie along a solution path. Some of these samples (purple) lie inside critical regions, such as narrow passages.}
    \label{fig:learning-sampling}
\end{figure}

One of the most intuitive ways to apply learning to \sbmps is to use it to improve the quality and efficiency of sampling.  Learned models can be used to produce samples that are more likely to be valid and/or occur in \emph{critical regions} such as narrow passages (Figure~\ref{fig:learning-sampling}).

\begin{mdframed}[hidealllines=true,backgroundcolor=red!20,frametitle=Categories of work on Sampling Sequences]
\begin{myitem}
    \item \textbf{Modeling the C-space} \citep{AT-2015,bb-2005,chamzas2019using,sutanto2020learning,9196771,kingston2019exploring}
    \item \textbf{Biasing samples along likely shortest paths} \citep{BN-2010,ZKB-2008,IHP-2018,kumar2019lego,Wang-2020,9197106,huh2018efficient}
\end{myitem}
\end{mdframed}

\subsection{Modeling the C-space}

This class of approaches learn a representation of a problem's $\cspace$ and query the learned model to generate samples that are likely to be in relevant regions, or to filter out samples that are likely to be invalid. 

One such method learns two probability distributions that model $\mathbb{C}_{free}$ and $\mathbb{C}_{obs}$ of a planning problem via a kernel density estimator function \citep{AT-2015}. The estimator approximates the likelihood that a drawn sample will be in $\mathbb{C}_{free}$ or $\mathbb{C}_{obs}$. The model is trained on previously drawn samples that have been collision-checked. During runtime, a Bayesian classifier is used to predict if a new sample lies in $\mathbb{C}_{free}$ or not. 

\textit{Model-predictive Motion Planning} \citep{bb-2005} uses locally-weighted regression to build a model that predicts if a given sample is in $\mathbb{C}_{free}$ or $\mathbb{C}_{obs}$. This approximate model is constructed incrementally as a solution to the motion planning problem is computed. The sampling strategy associated with the learned model adapts sampling densities in proportion to an area’s complexity, allowing it to explore new regions efficiently.

The \textit{Global-Local Sampler }(GL-Sampler) \citep{chamzas2019using} uses a learned database of local samplers to discover the connectivity of configuration space. It decomposes the workspace into a set of local primitives.  During the offline phase, it computes local samplers for these primitives and saves them to a database.  During the online phase, the GL-sampler maps primitives to local samplers from the database, and then uses these samplers to synthesize a global sampler.

Learned models of implicit manifold configuration space (IMACS) generate tangent spaces that represent manifolds \citep{kingston2019exploring}.  These tangent spaces provide piecewise-linear approximations of manifolds and can generate manifold samples and perform local planning on manifolds.

\textit{Equality Constraint Manifold Neural Network} (ECoMaNN) \citep{sutanto2020learning} is a neural network that models manifold constraints that are commonly found in tasks like grasping and manipulation. It takes a configuration as input and outputs a predicted value that indicates the distance from the manifold. ECoMaNN is trained to minimize the errors between the tangents and normals of the manifold and those predicted by the model. The training data is augmented using off-manifold points. ECoMaNN is used in combination with sequential motion planning methods to generate paths on these manifolds. 

Learned sampling  distributions have also extended sampling-based planners to unknown environments \citep{9196771}. The distributions are derived from both geometric representations (e.g. occupancy grids) and object-level representations (e.g. contextual cues), and trained using a dataset of example trajectories. A sampling-based planner samples vertices from the learned distribution and scores roadmap edges using a proposed objective function.

\subsection{Biasing samples along likely shortest paths}
 
A second class of approaches learn a model that identifies paths in the environment, then uses the model to bias sampling along likely paths. Such methods can be trained to identify higher quality paths quickly. 

Given paths produced by an expert, it is possible to use Learning from Demonstrations (LfD) to produce similar paths \citep{BN-2010}. This is accomplished by learning a non-parametric representation of the distribution of samples along the expert paths, modeled using an Infinite Gaussian Mixture Model (IGMM).  

Early work included learning a sample distribution function over a discretization of the workspace that is then used to bias sampling towards critical regions such as narrow passages \citep{ZKB-2008}.  Sampling is performed by obtaining an instance from the probability distribution, then randomly generating a sample from the distribution of samples given this instance.  The sample distribution function is obtained by reinforcement learning.  The resulting motion plans are then evaluated using a reward function, and the parameters of the sample distribution are adjusted towards the direction of the approximate gradient of the reward function.

Conditional Variational Autoencoders (CVAEs) have been used to effectively learn sample distributions \citep{IHP-2018}. CVAEs are advantageous because they can represent complex, high-dimensional distributions that can be conditioned on arbitrary problem inputs. The CVAE is trained from demonstrations (examples of successful motion plans, human demonstrations, etc.) on similarly structured environments. In the online operation of the motion planner, samples are generated via the latent space of the CVAE, conditioned on problem-specific variables, such as start, goal and obstacle information (for e.g., an obstacle occupancy map). Theoretical properties are maintained by also generating samples using an auxiliary (uniform) sampler over the state space.

While the above approach is shown to be effective on many problems, it can potentially fail in the face of complex obstacle configurations, or mismatch between training and testing environments. \textit{Leveraging Experience with Graph Oracles} (LEGO) \citep{kumar2019lego} addresses these issues by predicting samples that belong only to bottleneck regions, and which are spread out across diverse regions to maximize the likelihood of a feasible path existing.

The \textit{Neural RRT*} approach \citep{Wang-2020} uses a convolutional neural network (CNN) to bias sampling towards predicted optimal paths.  It formulates the problem of generating good samples as an image-to-image mapping problem where the network's input is an RGB image of the workspace and robot-specific attributes such as the step size and clearance.  The network outputs a probability density as a pixelation over the workspace image, where the probability associated with each pixel value indicates the likelihood that the pixel will contain the optimal path. The architecture is trained from a large dataset of optimal paths generated using an A* algorithm to minimize the cross-entropy between the predicted optimal-path probabilities and the ground truth. The predicted probability distribution is then used to generate samples to be used by an RRT* planner. In order to maintain probabilistic completeness, a proportion of samples are generated using uniform sampling.

The \textit{Critical PRM} \citep{9197106} identifies critical states along solution trajectories using graph-theoretic techniques, and builds a dataset of critical states and local environment features (such as an occupancy grid) for different planning problems. Then, in a new environment, a criticality prediction network is trained to predict the criticality of a new sample. In the proposed Critical PRM method, a proportion of states are sampled proportional to their criticality, while the rest are sampled uniformly. Critical PRMs are demonstrated to achieve up to three orders of magnitude improvement over uniform sampling, while preserving the guarantees and complexity of SBMP. 

\textit{Q-function Sampling RRT} (QS-RRT) \citep{huh2018efficient} learns a state-action-value function (Q-function) over $\cspace$, where a state corresponds to a node on the tree, and an action is an extension of the tree from that node. The approach makes use of a Radial Basis Feature (RBF) representation in $\cspace$ to improve the effectiveness of the Q-function learning. The Q-function is integrated using a softmax node selection procedure inside RRT, and the selection of the best tree expansion is performed using numerical optimization. 

\begin{mdframed}[hidealllines=true,backgroundcolor=blue!20,frametitle=Discussion on Sampling Sequences]
To use a learned sampling strategy for a class of motion planning problems, the variables that change across challenges must be clearly defined. For instance, in planning for a tabletop manipulation problem with a static set of obstacles whose poses may vary, the input to the learned model should be the start and goal configuration, as well as the object poses. 

For problems where it is straightforward to obtain a dataset of high quality paths (from previously computed solutions, human demonstration, etc.), the learned model can be trained to bias samples along likely shortest paths given the problem input. For problems with complex constraints like manipulation, it makes sense to use the learned model to predict valid samples.
   
As the dimensionality of the input increases, the ability of the learned model to predict good samples will depend both on the amount of data used to train the model and the approximation power of the model. As the size of the model grows, an additional consideration is the trade-off between speed and quality of produced samples: a random sampler may be able to produce lower quality samples at a much higher rate than a learned sampler.
\end{mdframed}

\section{Collision Checking}

\begin{figure}[t]
    \centering
    \includegraphics[width=.99\textwidth]{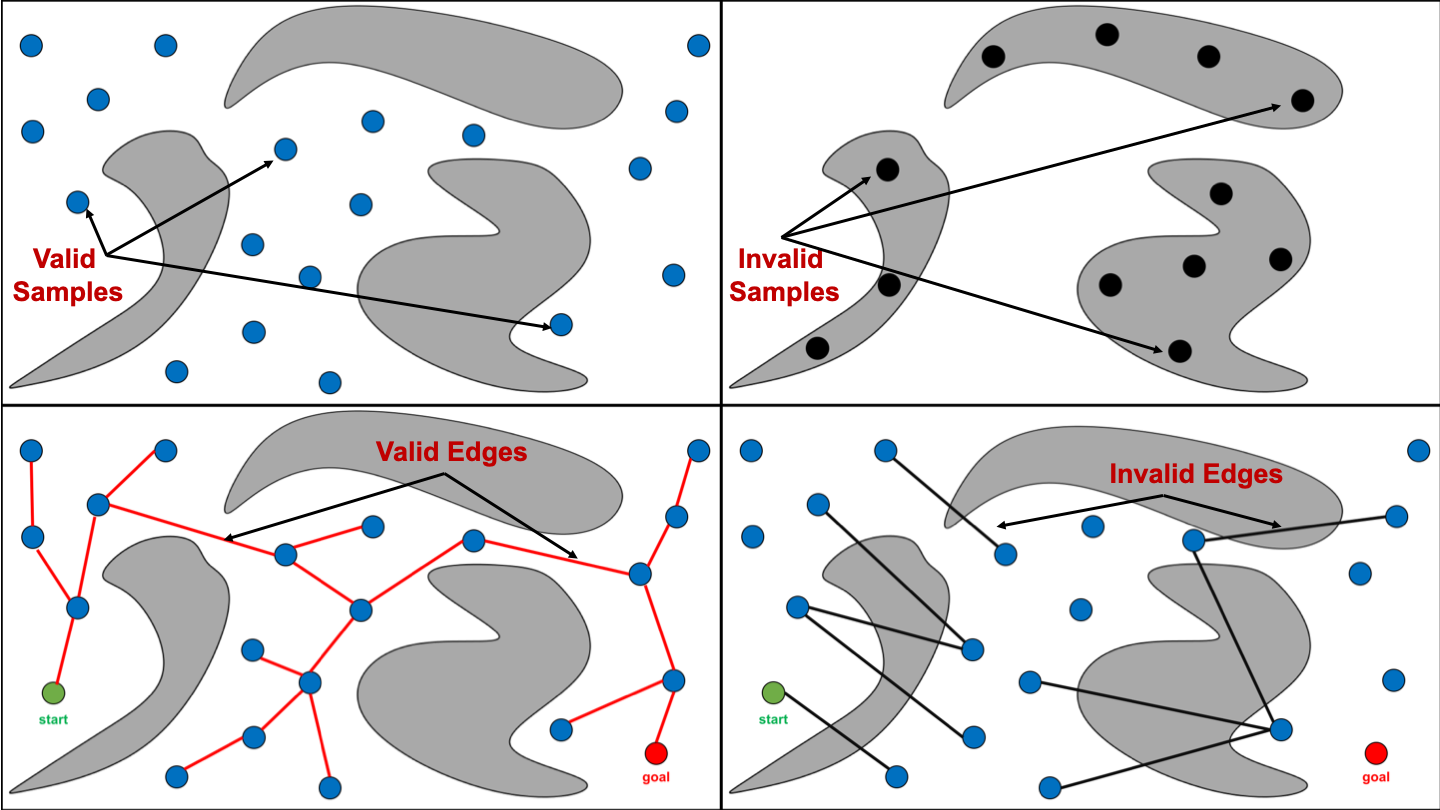}
    \caption{Given a representation of obstacles, a learned collision detection model provides a prediction that can be used to validate (at least) one of the following conditions: \textbf{(Top-left)} Samples in $\mathbb{C}_{free}$. \textbf{(Top-right)} Samples in $\mathbb{C}_{obs}$. \textbf{(Bottom-left)} Edges entirely in $\mathbb{C}_{free}$. \textbf{(Bottom-right)} Edges intersecting with $\mathbb{C}_{obs}$.}
    \label{fig:learning-cc}
\end{figure}

Collision detection is often the most computationally expensive step of solving a motion planning problem \citep{Kleinbort2020}. It is therefore a promising direction to use machine learning to reduce the cost of this step (Figure~\ref{fig:learning-cc}). In order to avoid collisions in the real world, prediction errors in the form of false positives and false negatives must be suitably mitigated. 

A practical strategy to minimize the number of collision checks while retaining theoretical properties of the motion planner is the idea of \textit{lazy collision checking} \citep{bohlin2000path,hauser2015lazy,mandalika2019generalized}. Under this strategy, edges are not immediately checked for collision, but rather are checked only when a candidate path to the goal is found. Feasible edges are marked and infeasible ones are deleted. The process repeats until a feasible path to the goal is found. This strategy avoids checking the vast majority of edges that have no chance of being on an optimal path.

\begin{mdframed}[hidealllines=true,backgroundcolor=red!20,frametitle=Categories of work on Collision Checking]
\begin{myitem}
    \item \textbf{Identifying samples that are guaranteed to be valid} \citep{BOKF-2013,BOF-2013}
    \item \textbf{Using a learned model in place of a collision detector} \citep{bb-2005,huh2016collision,Das_2020,kew2019neural,chenning2021collision}
    \item \textbf{Determining the order in which to collision check nodes or edges} \citep{PCM-2013,bhardwaj2019leveraging, Choudhury-2017-104364, Choudhury_2018, hou2020posterior}
\end{myitem}
\end{mdframed}


\subsection{Identifying samples that are guaranteed to be valid}

This area of work uses learned models to identify samples that are guaranteed to be valid, eliminating the need to test them using an expensive collision checker. The \textit{Safety Certificate} approach \citep{BOF-2013,BOKF-2013} proposes an adaptive sampling distribution that provably converges to a uniform distribution.  When a sample is generated and collision-checked traditionally, this method stores a lower bound on the sample’s distance to the nearest obstacle. This distance, referred to as the \emph{safety certificate}, defines a region of the search space that is guaranteed to be collision-free.  Subsequent samples that are generated within a safety certificate are guaranteed to be valid and do not need to be collision-checked.  As the number of samples goes to infinity the safety certificates asymptotically cover the entire search space and the amortized cost of collision checking becomes negligible with respect to the overall running time of the algorithm.


\subsection{Using a learned model in place of a collision detector} 

The most common approach to learned collision checking is to replace a traditional collision checker with a learned model.

Model-based Motion Planning \citep{bb-2005} also helps perform predictive edge validation rather then calling a dedicated collision detector. As discussed in Section \ref{sec:sampling}, this method uses locally weighted regression to build an approximate model of $\cspace$.  This model can be used to predict if unsampled regions of $\cspace$ are valid or invalid, and if edges are valid.  

Gaussian mixture models (GMMs) have been used for collision checking in RRTs \citep{huh2016collision}. The GMM is trained online during execution of the RRT using Incremental Expectation Maximization (EM). The training examples are generated by a traditional collision checker. By using biased sampling from the learned GMM distribution, the number of collision checks required for future iterations is shown experimentally to be reduced.

The \emph{Fastron-RRT} \citep{Das_2020} uses a kernel perceptron to create a model of $\cspace$ that can be used as a proxy kinematic-based collision detector.  The kernel perceptron generates a set of support points that form a separating hyperplane between two classes (in this case, $\mathbb{C}_{free}$ and $\mathbb{C}_{obs}$).  Samples are collision-checked by querying which side of the hyperplane they lie on.  \textit{Fastron-RRT} can also operate under an active learning setting that updates the model's hyperplane in order to reflect changes in the environment. 

\emph{ClearanceNet-RRT} \citep{kew2019neural} is a highly parallelizable extension of Fastron-RRT that uses a neural network (ClearenceNet) to predict obstacle clearance in order to obtain a heuristic for collision checking. The neural network obtains as input the poses of the robot and the obstacles in the environment, and outputs the distance to the nearest obstacle. This allows for the construction of an RRT where collision-checking can be deferred. Once a path has been found, ClearanceNet-RRT fixes errors in the path by performing gradient descent based path repair to move the path away from obstacles. 

A recent approach to learned collision checking employs graph neural networks (GNNs) to reduce the number of collision checks on a Random Geometric Graph (RGG) \citep{chenning2021collision}. Given the vertices and edges of the graph, as well as problem-specific information, such as obstacle information and goal location, an attention-based neural network outputs vertex and edge embeddings, which are then used by the GNN. The GNN consists of a path exploration component that iteratively predicts collision-free edges to prioritize their exploration, and a path smoothing component that optimizes the obtained paths.

\subsection{Determining the order in which to collision check nodes/edges}

Another class of approaches use a learned model to determine the order in which nodes/edges are tested (collision checked) in order to more efficiently identify a valid path.

Instance-based learning can be used to store previous local planning queries and their outcomes (i.e. whether they were in collision or collision-free) \citep{PCM-2013}. When a new sample is generated, this method identifies the $k$ nearest neighbors (or approximate nearest neighbors) from the nodes stored in a hash table. Given the neighbor set, the probability that the node is in collision is computed using a softmax function. This probability is then used as part of a collision filter for local planning queries.  It is also used as part of a method for exploring $\cspace$ around an existing sample in order to identify new samples for applications like RRT expansion.  Finally, it is used to defer collision detection to query time in order to perform collision detection in a lazy manner.

\textit{Search Though Oracle Learning and Laziness} (STROLL) \citep{bhardwaj2019leveraging} uses a neural network to determine which edges to collision-check expensively when performing a lazy search on a roadmap. It formulates the problem of edge selection as a Markov Decision Process (MDP) over the state of the search problem, and uses Q-learning to solve this MDP. 

The problem of path validity prediction can be formulated as a Decision Region Determination (DRD) problem \citep{Choudhury-2017-104364, Choudhury_2018}. This method first creates a training set from a sampled set of possible \textit{worlds} (environments), that form a set of hypotheses.  Then, it generates a roadmap without testing the edges and obtains a set of possible paths.  Each path is valid in some of the sampled worlds and invalid in others.  The set of worlds in which the path is valid forms the decision region for that path.  This method then uses the \textit{DIRECT} algorithm to obtain a decision tree that indicates what edges should be tested in order to prune inconsistent worlds and identity if a path is valid.

\textit{Posterior Sampling for Motion Planning} (PSMP) \citep{hou2020posterior} formulates anytime search on graphs as an instance of Bayesian Reinforcement Learning (Bayesian RL). Unlike prior work, PSMP aims for anytime performance by leveraging learned posteriors on edge collisions to quickly discover an initial feasible path and progressively yield shorter paths.

\begin{mdframed}[hidealllines=true,backgroundcolor=blue!20,frametitle=Discussion on Collision Checking]
The most straightforward way of applying a learned collision detector is to model it as a classification problem. For a static environment, an offline dataset of robot configurations in $\mathbb{C}_{free}$ and $\mathbb{C}_{obs}$ can be collected to train an appropriate machine learning model that predicts if an input configuration is in collision. It can also be beneficial to use a learned model that outputs a \textit{confidence score} about its prediction (e.g., a Bayesian model that outputs class probabilities). For predictions that have a confidence score below a fixed threshold the planner can query a more expensive, traditional collision checker.

If traditional collision checking, including with lazy evaluation, proves to be relatively expensive, it makes sense to use a learned model to instead determine the order in which to test nodes/edges. 
\end{mdframed}


\section{Distance Metrics, Heuristics and Constraints}
\label{sec:learned-distance}
In SBMP, distance metrics are used to estimate the viability of connecting samples and as a cost function for optimization. A common choice of metric is the Euclidean distance, under the intuition that the further the robot has to travel between states, the more likely that it will encounter an invalid state along its path. However, this metric does not take into consideration the topology of the environment, the complex motions performed by the rigid bodies of the robot, or the robot dynamics. For applications where a more accurate distance metric may be available, it may be computationally prohibitive to compute. Moreover, for many applications it is not immediately obvious what an accurate distance metric would look like. 

\begin{figure}[t]
  \begin{center}
    \includegraphics[width=\textwidth]{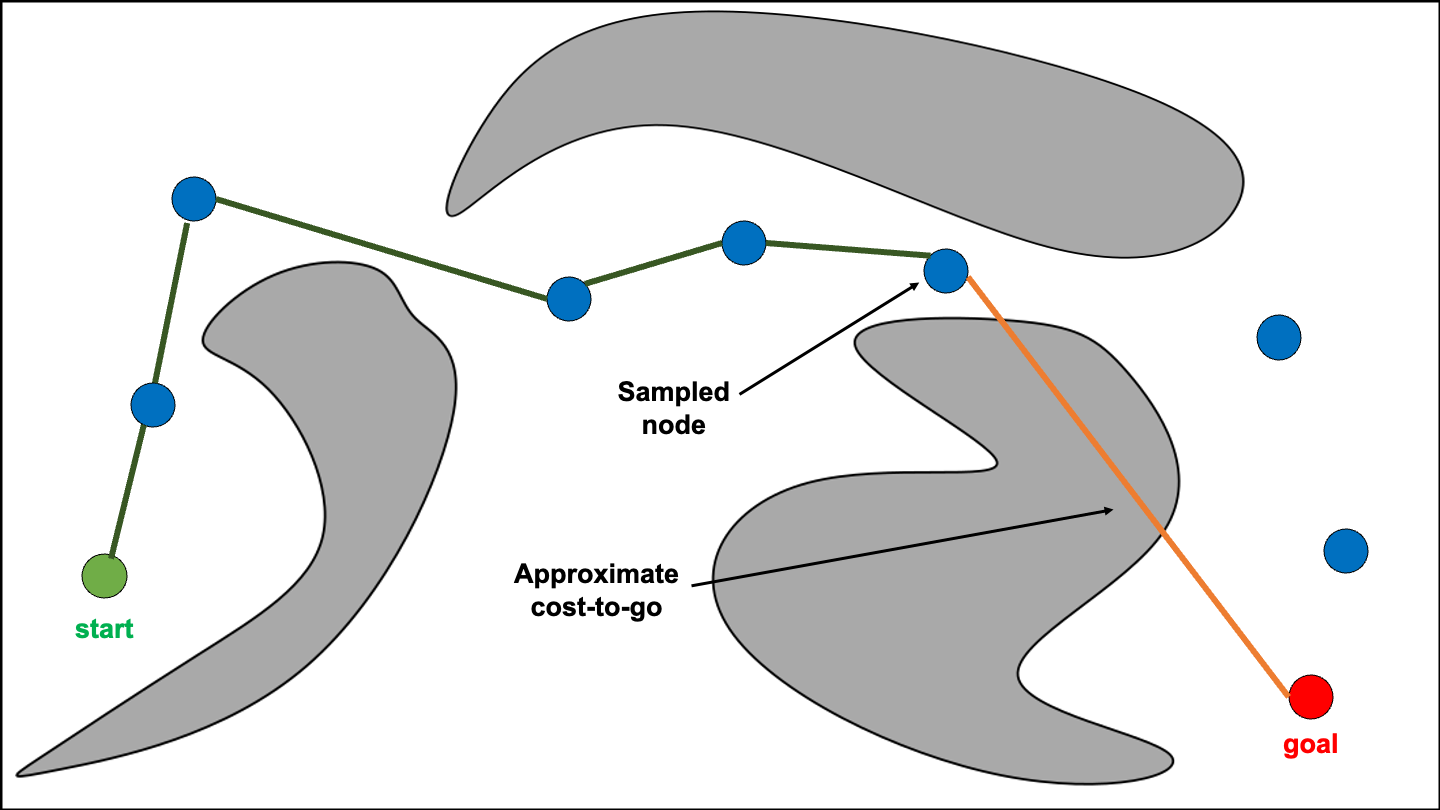}
    \caption{The \emph{cost-to-go} is a heuristic estimate of the cost to move the robot from its current configuration to the goal.  In this example, the Euclidean distance of the sample to the goal (orange line) is used to approximate the cost-to-go.}
    \label{fig:learning-cost-to-go}
  \end{center}
\end{figure}

Many sampling-based motion planners also use an estimate of the \textit{cost-to-go} from a node to the goal in order to bias the selection of nodes that are closer to the goal, and as a heuristic to speed up the search for the best solution. If the cost-to-go metric used is \textit{admissible}, then it can be added to the cost of getting from the start node (the \textit{cost-to-come}) to obtain a lower bound on the possible cost of paths that go through the node (Figure~\ref{fig:learning-cost-to-go}).  This can be used to give preference to expanding nodes that may yield shorter paths to the goal, while pruning nodes that cannot.

\begin{mdframed}[hidealllines=true,backgroundcolor=red!20,frametitle={Categories of work on Distance Metrics, Heuristics \& Constraints}]
\begin{myitem} 
    \item \textbf{Approximating distance metrics\\} \citep{cfst-2018,6942569,pa-2015,LB-2011,kvea-2018}
    \item \textbf{Approximating cost-to-go values} \citep{AT-2015,PhillipsGrafflin2015ReproducingEM,YA-2017,huh2020cost}
    \item {\bf Planning for systems with dynamics} \citep{KP-2007,LCLX-2018}
    \item \textbf{Modelling the effect of a heuristic} \citep{HBL-2005,Bhardwaj-2017-101433,yuanknowledge,wells2019learning,kim2019learning, zhao2020plrc} 
\end{myitem}
\end{mdframed}

\subsection{Approximating the distance metric}

A straightforward approach is to use a machine learning model to learn an approximation of a complex distance metric. The learned model can then be used in place of a computationally expensive distance computation step in the planning phase. 

For instance, the \textit{swept volume} is the volume in the workspace, that the local planner sweeps over while passing between two configurations. It is generally considered to be an ideal distance metric, however it is very expensive to compute online. Deep neural networks can be used to estimate the swept volume between two points \citep{cfst-2018}. The network is trained in an obstacle-free environment from examples of pairs of configurations and the swept volume between them. The trained network, alongside a hierarchical neighborhood search procedure, is shown to be able to approximate swept-volume distance effectively and efficiently.

Parametric models like locally-weighted projection regression can be used to approximate distances \citep{6942569}. The model is trained by approximating the optimal cost between pairs of states using an iterative Linear Quadratic Regulator (iLQR). The learned model approximates the original cost with a reasonable tolerance and gives a speed-up of a factor of 1000 over computing the actual cost.
It is also possible to use a simpler non-linear parametric model to approximate distances \citep{pa-2015}. The model is trained using a novel steering function that solves the two-point boundary value problem (BVP) between two configurations of a wheeled mobile robot. 

The distance between two states can also be approximated by constructing a densely sampled roadmap offline and querying the roadmap during planning time \citep{LB-2011}. The graph is generated by sampling a large number of states in an obstacle-free environment and then applying controls to the states. Two states ($s_i,s_j$) are considered to be connected if there is a control that extends the state $s_i$ to a state that is close to $s_j$. To reduce the online cost, the sampled states are mapped into a higher-dimensional Euclidean space through multi-dimensional scaling (MDS) that retains the relative distances represented by the sampled graph.

Orthogonal to the other methods discussed above, the \textit{Distance-Aware Dynamic Roadmap} (DA-DRM) \citep{kvea-2018} learns a voxel distance grid which is updated using perceptual information from a humanoid robot’s perception system.  The learned distance information is use to estimate costs during roadmap search, and for path smoothing.  

\subsection{Approximating the cost-to-go value}

Informed \sbmps \ typically exploit the structure of the planning problem via an approximate cost-to-go value (or a \textit{heuristic}) to improve the practical efficiency of the search process \citep{gammell2021asymptotically}. It may thus be beneficial to learn a function that approximates the problem-specific cost-to-go.

Similar to early work in approximating distance metrics, early work models the problem of learning an approximate cost-to-go as a supervised regression problem \citep{AT-2015}. It uses a training dataset where each data point consists of a randomly drawn point along with its \textit{locally minimum cost-to-come estimate} (lmc value).  Distance-weighted regression is used to fit a surface from points in the training set and to estimate estimate the cost-to-go of new points.  See Section \ref{sec:sampling} for additional information on the proposed method.

For problems where the cost metric may be difficult to mathematically represent (like robot surgery), it can be learned from demonstrations through active learning and Inverse Optimal Control (IOC) \citep{PhillipsGrafflin2015ReproducingEM}. Similarly, \textit{Demonstration-Guided Motion Planning} (DGMP) \citep{YA-2017} learns a cost function that encodes task constraints for household tasks (for e.g., a full glass of water must be held upright) from expert demonstrations of the task. The learned cost function extracts time-dependent task constraints by learning low variance aspects of the demonstrations, which are correlated with the task constraints. The learned cost metric is optimized using \textit{Multi-Component Rapidly-Exploring Roadmaps} (MC-RRM), a SBMP technique. 

A neural network architecture, {\tt c2g-hof} \citep{huh2020cost}, takes as input a point cloud representation of the workspace and a goal configuration to output a cost-to-go estimate. It is trained on a dataset of costs between every pair of configurations in randomly generated workspaces. During execution, the gradient of the output cost-to-go is followed to yield continuous, collision-free trajectories.

\subsection{Planning for systems with dynamics}

Planning with \textit{viability filtering} \citep{KP-2007} learns a viability model of an agent’s \emph{perceptual space} that is used to direct planning.  This method uses a virtual range finder sensor to model the robot's perception at a given state. It trains a viability classifier on a large set of motions along with their associated virtual perceptions that can be obtained from previous solutions or from simulation.  This method uses the classifier online to limit the planner's search to motions that lead to viable regions.  It improves planning efficiency by preventing the planner from exploring nonviable regions, which will unavoidably lead to failure.

Near-Optimal RRT (NoD-RRT) \citep{LCLX-2018} uses a neural network to predict the cost between two given states considering nonlinear constraints. The neural network is trained on examples where the two point BVP is solved between pairs of states.  The authors show theoretically that NoD-RRT is asymptotically optimal.

\subsection{Modelling the effect of a heuristic} 

In some cases, it may be easier to model the \textit{effect} of using a good heuristic, such as the order in which nodes must be expanded during search. 

In applications like rock climbing where multiple queries (locomotion, manipulation, contacts) must be processed to solve a specific planning problem, it is imperative to decide quickly if a query is solvable or not. Towards this objective, it is possible to train a classifier \citep{HBL-2005} to predict if a query is solvable.  The classifier is trained using data obtained from many iterations of Basic-MSP (multi-step planning) in randomly generated terrains.  This classifier can then be integrated with a multi-step planning algorithm to avoid spending computational effort in solving infeasible queries.

\textit{Search As Imitation Learning} (SAIL) \citep{Bhardwaj-2017-101433} uses a heuristic policy that is trained to imitate an oracle that has perfect knowledge of the world and is capable of making decisions that minimize the required search.  SAIL posits that the manner in which a tree-based planner explores an environment can be seen as a wave-front and that it is desirable to keep the wave-front as small as possible while focusing the search towards the direction of the goal. It formulates the problem of selecting nodes for expansion as a sequential decision making problem in which information extracted from the wavefront is used to decide what nodes to expand. It then uses dynamic programming to compute the heuristic values for all states to determine the order of exploration.

\textit{Piecewise Linear Regression Complex} (PLRC*) \citep{zhao2020plrc} is a data structure that approximates the costs of crossing local regions of a configuration space using piecewise linear regression. It is constructed by decomposing $\cspace$ into a set of cells. For each cell, pairs of points are sampled along the boundary, and the distance between them is calculated and stored. Then, this data is used to build a regression model that can return the costs between boundary points of every cell. For simple problems, the memory required to store this representation compares favorably to that required for standard discrete vertex-and-edge models, while achieving similar path quality.

There has been recent interest in methods that encode prior information about the planning problem in the form of a sparse data structure. \cite{yuanknowledge} constructs a topological feature tree and computes a heuristic path on the tree. The computed path is then integrated with an \sbmp\ to improve the success rate in environments with dynamic obstacles. 

Learned models have also been used to improve the efficiency of Task and Motion Planning by learning a classifier for feasible motions, and using the output of this classifier as a heuristic for guiding the search \citep{wells2019learning}. The classifier is trained on minimal exemplar scenes. On more complex scenarios, principled approximations are applied so as to minimize the effect of errors. Even in the presence of classification errors, properly biasing the heuristic ensures that the search is still orders of magnitude faster than an uninformed search.

An algorithm for learning to guide a planner \citep{kim2019learning} proposes learning to predict constraints on the solution. A new type of problem-instance representation called \textit{score space} is proposed. Its main advantage is that it relays information about the similarity of problem instances. To transfer knowledge from past experience, the algorithm uses the correlation information in the problem score space representation of a given instance.  This limits planning to the constrained subspace, which reduces the size of the space that needs to be searched.

\begin{mdframed}[hidealllines=true,backgroundcolor=blue!20,frametitle=Discussion on Distance Metrics \& Heuristics]
Learned models, such as those based on neural networks, can be used to approximate a more expensive to compute distance / cost-to-go metric, if a considerably large dataset of pairs of states and the true distance / cost-to-go between them can be collected. Access to the aforementioned offline dataset is not possible for some systems because it is not clear what a good distance / cost-to-go metric would look like. Approximating the distance metric computation using a learned model opens up interesting avenues for speeding up the nearest neighbor search component as well. There is a caveat, however. Due to the inherent bias-variance trade-off in any machine learning model, it is possible for the predicted distance / cost-to-go to have some error, which must be taken into consideration. 

It is also possible to use learning to simulate the \textit{effect} of using a good heuristic, that biases the sampling of nodes to find higher-quality solutions more quickly - similar to some of the models discussed in Section~\ref{sec:sampling}. These methods, however, are not directly applicable to systems with constraints on their motion (such as dynamics), or where the cost function being optimized encodes task-specific attributes (such as grasping). 
\end{mdframed}

\section{Steering Functions and Local Planners}
One of the requirements of many state-of-the-art \sbmp s is the availability of a local planner (or a \textit{steering function}) that connects two states of the system being planned for.  For some problems, such as those with constraints on the robot dynamics, a steering function corresponds to a boundary value problem (BVP) solver, which may not be available, or may be computationally expensive to query. This has led to the development of learned models that can be used in place of local planners.

\begin{mdframed}[hidealllines=true,backgroundcolor=red!20,frametitle=Categories of work on Steering Functions \& Local Planners]
\begin{myitem}
    \item \textbf{Learned steering functions for kinodynamic systems } \citep{FPCFT-2017,faust-acta-14,forftfd-2018,Chiang_2019,learned_goal_reaching_controllers,zheng2021sampling}
\end{myitem}
\end{mdframed}

Reinforcement learning offers many promising avenues to learn a local planner (or a policy). It has been used to enable aerial robots to quickly navigate an environment and deliver payloads while complying with the dynamic constraints introduced by these loads \citep{FPCFT-2017, faust-acta-14}. The proposed method learns a policy for the system, that minimizes the residual oscillations in the load.  The large size of the action space associated with planning for aerial vehicles makes direct policy learning impractical for this system; so the policy is trained on a simplified problem space. During the planning stage, a PRM is used to generate collision-free paths.  The learned policy is then used to transform these paths into trajectories that satisfy the constraints imposed by the suspended load. 

\textit{PRM-RL} \citep{forftfd-2018} is a long-distance navigation method that uses reinforcement learning to obtain a policy for point-to-point navigation. The policy is first trained in a simple environment where it learns dynamics of the system, along with any task constraints and sensor noise.  The trained policy is then used during edge connection to obtain paths between candidate neighbors.  Neighbors are connected if the learned policy can consistently find a path between them. Similar to PRM-RL, \textit{RL-RRT} \citep{Chiang_2019} uses a learned obstacle-avoidance policy to perform local planning between two states, along with a learned reachability estimator to bias tree growth. The reachability estimator is trained via supervised learning to predict the time required for the obstacle-avoidance policy to reach a state given the obstacles that are present. 

Recent work \citep{learned_goal_reaching_controllers} has proposed training a goal-reaching controller in an obstacle-free environment and applying this controller for planning in environments with obstacles. The node expansion procedure makes use of an informed local goal generation strategy for input to the learned controller. The expansion procedure is integrated inside the AO DIRT \citep{LB-DIRT,ML4KP} kinodynamic planner for improving success rate and solution quality in vehicular navigation problems.

If optimal trajectories between two states can be computed offline, a supervised learning controller can be trained to steer the system from points in an initial state set to points in a final goal set. Learning-based kinodynamic RRT* \citep{zheng2021sampling} integrates this learned controller along with a learned cost-to-go metric.

\begin{mdframed}[hidealllines=true,backgroundcolor=blue!20,frametitle=Discussion on Steering Functions and Local Planners]
In order to improve the efficiency of sampling-based kinodynamic planning via a learned model, it is possible to either learn a function that predicts the cost-to-go between two states, or a local planner than can viably connect two states, or both. Recent advances in deep reinforcement learning have focused on learning value functions and policies for a variety of different robotics tasks, but require a large dataset of interactions with the environment. It is a promising area of research to integrate these RL advances into SBMPs, which allow for planning over longer horizons, while maintaining theoretical properties and sample efficiency considerations.
\end{mdframed}

\edgar{ 
}



\section{Adaptive Selection of Primitives and Meta-Reasoning} 
\label{sec:adaptive_planning}
Many planning algorithms have been developed that use different combinations of sampling methods, neighborhood finders, expansion methods and connection strategies. It is often difficult to ascertain what combination of methods is best for a particular problem. One way of addressing this issue is to use learning to predict what methods or combinations of methods work best for a new planning problem. 
In some cases, it is useful to determine when to stop computational effort for a given planning problem. This form of reasoning is known as \textit{meta-reasoning}, and recent work has proposed machine learning-based solutions.

\begin{mdframed}[hidealllines=true,backgroundcolor=red!20,frametitle=Categories on Adaptive Selection \& Meta-Reasoning]
\begin{myitem} 
    \item \textbf{Planner selection} \citep{mmtpra-2005,Choudhury-2015}
    \item \textbf{Connection strategy selection} \citep{EJTA-2013,Uwacu2016TheIO}
    \item \textbf{Sampler selection} \citep{HSS-2005}
    \item \textbf{Paths \& parameter selection} \citep{chamzas2020learning, moll2021hyperplan}
    \item \textbf{Meta-reasoning} \citep{li2021learning, sung2021learning}
\end{myitem}
\end{mdframed}

\subsection{Planner Selection}

One adaptive selection method uses hand-crafted features in the environment to determine which combination of primitive methods to use \citep{mmtpra-2005}. It first partitions the environment by recursively dividing $\cspace$  until it obtains regions that are classified as homogeneous according to a set of features, which are obtained by constructing a simple PRM in each region and collecting statistics such as free node ratio, number of connected components, etc. Then, it uses a decision tree classifier which labels each region as free, cluttered, narrow passage or non-homogeneous based on the features observed in that region.  The decision tree is constructed using feature data that is collected from known examples of free, cluttered, narrow passage, and non-homogeneous environments. The planner that is selected for each region type is specified by the user (for e.g. PRM in the free regions and OBPRM in cluttered, narrow passage, and non-homogeneous environments). 

The \textit{Planner Ensemble} \citep{Choudhury-2015} learns a mapping from applications to appropriate methods. From a set of complementary planners (including both sampling-based planners and trajectory optimization methods), it selects which planner to use for a specific application by learning priors on the planners' performance. The priors are then used to estimate the probability that each method will be able to solve a new problem in order to select the ensemble of planners that has the maximum likelihood of finding a feasible solution.  The model is trained on a dataset consisting of 10k planning problems that are generated by taking permutations and combinations of various primitive scenarios.

\subsection{Neighborhood Connection Strategy Selection}

The \textit{Adaptive Neighbor Connection} (ANC) strategy \citep{EJTA-2013} uses a learned model to select from a set of neighborhood connection strategies based on their cost and how well they perform.  A neighborhood connection strategy consists of a candidate neighbor selection method (e.g. $k$-closest, $k$-Rand, etc.) and a distance metric (e.g. Euclidean, swept volume). ANC maintains a selection probability for each method, which is initially the same for all methods.  ANC selects a method according to the probability distribution, and after the connection step, it applies a reward or penalty to the selected method based on how many connections were successful and the number of collision detection calls made.  It then updates the selection probability of the methods based on this penalty/reward. An extension of ANC \citep{ETA-2015} uses a \emph{local} learning method that selects appropriate connection methods for roadmap construction.  As with ANC, this method dynamically divides the environment into regions.  It then selects what connection policy to use in each region using an adaptive probability distribution.  Extensions of this work have shown that introducing randomness into neighbor selection during training improves learning efficiency \citep{Uwacu2016TheIO}.

\subsection{Sampler Selection}

Another approach uses machine learning to adaptively combine multiple samplers (such as a Gaussian or Uniform sampler) by observing the performance of each sampler, and selecting the samplers that perform well more frequently \citep{HSS-2005}. This method maintains a probability for each sampler based on its cost and performance. To compute this probability, it maintains a weight for each sampler that captures its past performance. In order to ensure that all samplers have a reasonable chance of being tried, the weights for each sampler are initially assigned to be equal.   These weights are then used to compute a \emph{cost-insensitive} probability, which is scaled by the inverse of the cost of that sample to obtain a selection probability for each sampler.  At each timestep, the method selects a sampler based on this probability. 

\subsection{Paths \& Parameter selection} 
\textit{SPARK} and \textit{FLAME} 
\citep{chamzas2020learning} are experience-based frameworks for complex manipulators in 3D environments.  Both \textit{SPARK} and \textit{FLAME} build an experience database of local primitives, which are created through a decomposition of the environment to capture local workspace features.  Each primitive is associated with a local sampler, which, given previous experiences, is able to generate samples in critical regions of the primitive.   This method solves motion queries by retrieving similar primitives from the database and combining their local samplers. A follow-up work, \textit{FIRE} \citep{chamzas2022learning} extracts local representations of various planning problems to learn the similarity function over them.

\textit{HyperPlan} \citep{moll2021hyperplan} formulates the problem of motion planning algorithm selection and parameter optimization as a \textit{hyperparameter optimization} problem. It defines a loss function that captures the planning speed, combined planning and execution time, as well as the convergence to optimal solutions. Bayesian Optimization can then be used to optimize this loss function.

\textit{ALEF} \citep{kingston2021using} uses experience from similar planning queries to efficiently solve constrained motion planning for multi-modal problems, such as the ones that arise during manipulation when an object is grasped, or the end effector is constrained to move along a straight-line path. ALEF builds a sparse roadmap within an augmented, manifold-constrained state space which aggregates experience
gathered from different single-mode problems within a mode
family. Upon a new query, paths from this roadmap are retrieved and used to bias sampling in an \sbmp.

\subsection{Meta-Reasoning}

Given the intuition that search trees rooted at the start and goal of a planning problem form two separate classes in $\cspace$, recent work has proposed learning a manifold that separates these classes \citep{li2021learning}. Points are then sampled on the manifold, and a closed polytope is constructed to approximate it. Under some assumptions, the output is either an infeasibility proof of the planning problem, or a motion plan that connects the start and goal states.

Data-driven learning methods for \emph{meta-reasoning} have been used to model the relationship between solution quality and planning time in order to determine when to stop planning \citep{sung2021learning}.  Data for this method is generated by running an anytime planner on training problems for a long enough time to obtain an optimal solution with a high probability.  A model-based meta-reasoning method is proposed, which formulates the problem as a MDP, along with a model-free variation, which uses supervised learning in the form of a recurrent neural network (RNN).

\begin{mdframed}[hidealllines=true,backgroundcolor=blue!20,frametitle=Discussion on Adaptive Selection \& Meta-Reasoning]
The application of machine learning tools to adaptively select sampling-based motion planning primitives offers many promising avenues for exploration. Existing methods estimate the utility of different primitives or their combination by constructing a probability distribution that reflects their relative utility. These distributions are conditioned on features extracted from the environment, which can either be handcrafted or learned through a more expressive model. It is also interesting to consider learning the parameters of this distribution in a reinforcement learning (RL) setting, rather than a supervised setting. 
        
Another set of approaches accommodate heterogeneous problems by dividing them into \textit{regions}, and maintaining a separate probability distribution for each region based on the performance of different primitives in that region. Given the advances made in the expressive power of models like deep neural networks, it would be interesting to see if the classification of these regions can be learned automatically from prior planning experience, rather than through handcrafted features. 

\end{mdframed}
 
        

\chapter{Learning-based Pipelines}
\label{sec:integrated}
This chapter looks at machine learning models that learn from prior experience to help solve a sampling-based motion problem, but do not explicitly focus on learning a single primitive.

\begin{mdframed}[hidealllines=true,backgroundcolor=red!20,frametitle=Categories of work on Learning-based Pipelines]
\begin{myitem}
    \item \textbf{Retrieve and repair methods} \citep{BAG-2012,CSMOC-2014}
    \item \textbf{Neural motion planning} \citep{qureshi2019motion,JurgensonT19,strudel2020learning,dynamic-mpnet,li2021mpc, 9501956, 9143433}
    \item \textbf{Learning a lower dimensional planning space} \citep{ichter2019robot, chen2019learning, ichter2021broadlyexploring}
\end{myitem}
\end{mdframed}

\section{Retrieve and Repair Methods}

\begin{figure}[h!]
    \centering
    \includegraphics[width=.47\textwidth]{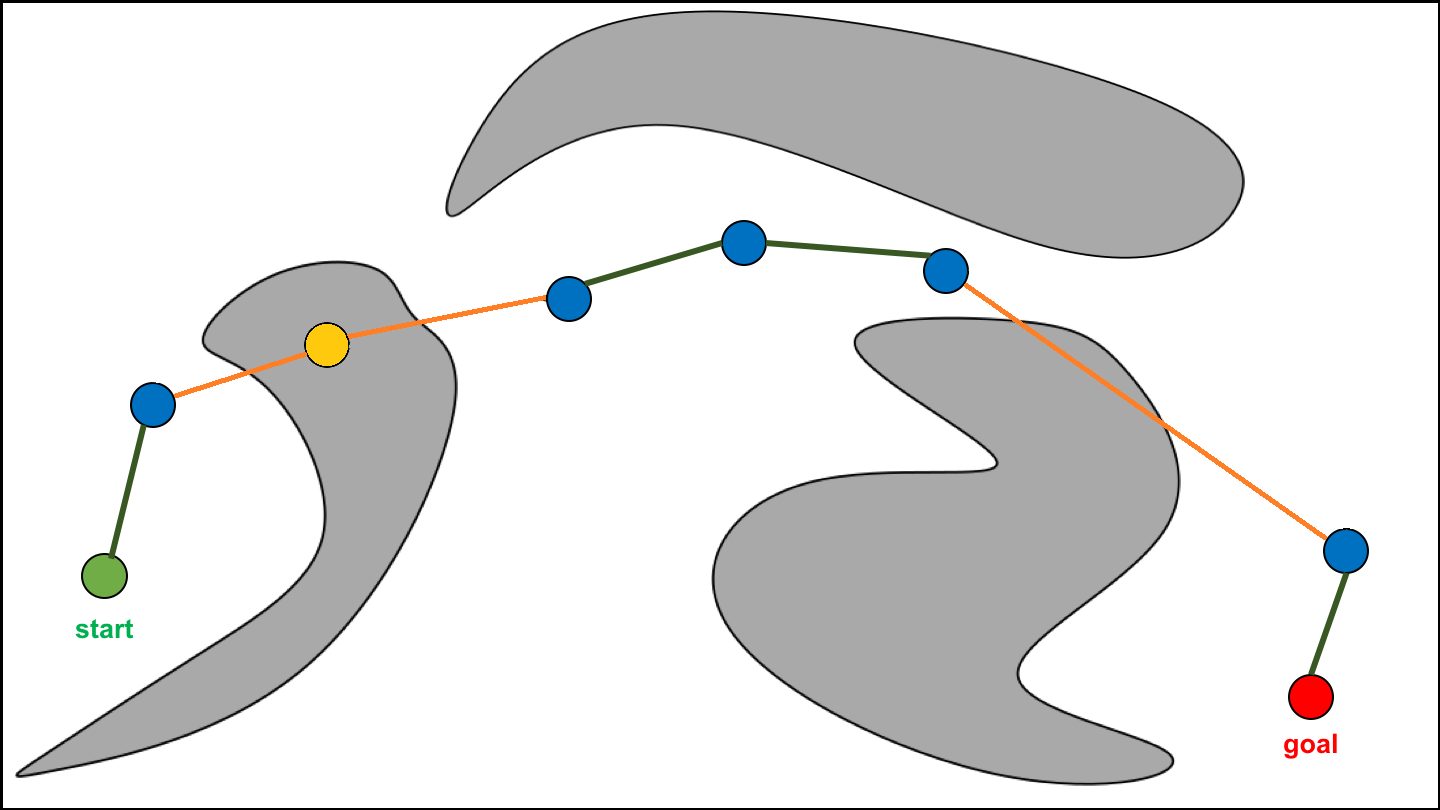}
    \includegraphics[width=.47\textwidth]{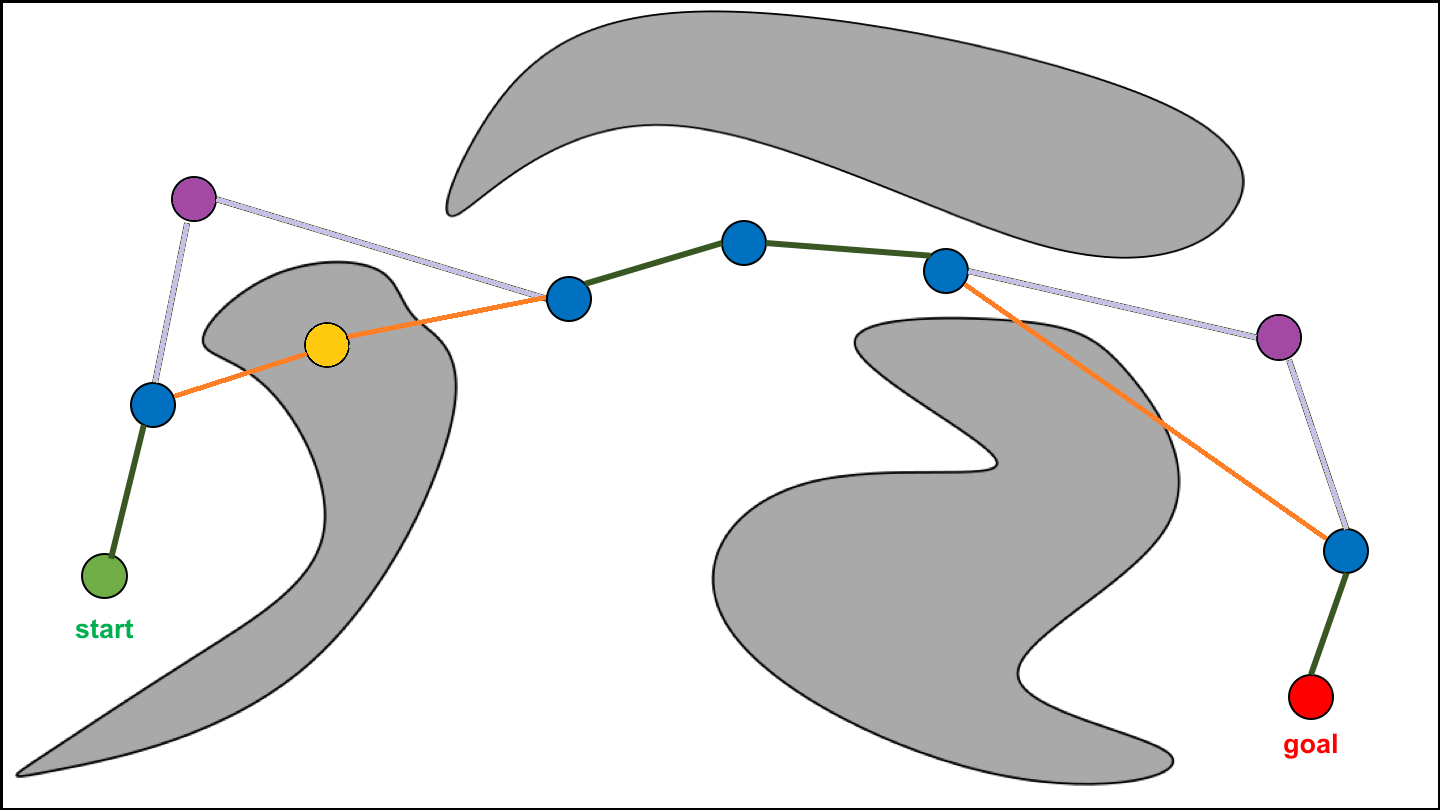}
    \caption{Retrieve and repair methods (left) retrieve a path from a library.  The retrieved path may have parts that are invalid (orange).  (right) These methods then repair the invalid portions by adding new nodes and edges (purple)}
    \label{fig:repair_and_retrive}
\end{figure}

This class of models use a database of paths and solve queries by retrieving these paths and adjusting them to generate solutions to new problems (Figure \ref{fig:repair_and_retrive}). The \emph{Lightning} framework \citep{BAG-2012} incorporates a \emph{Retrieve-Repair} module that retrieves a path from a library stored on the cloud that includes experience (e.g. paths) collected from multiple robots.  \emph{Lightning} selects a path from the library using two heuristics: one that quickly selects a set of candidate paths using the distance between their endpoints and the start configuration, and one that selects the path with the least constraint violation among these candidates.  

Similarly, the \emph{Thunder} framework \citep{CSMOC-2014} is an experience-based planner that stores paths obtained by probabilistic sampling in a SPArse Roadmap Spanner (SPARS). To solve a new problem,  it searches the database for a solution that was able to solve a similar problem and uses this solution as a basis to solve the new problem.  It identifies parts of the solution that are no longer feasible given the new problem and ``repairs'' them to form a valid solution.  This approach is well suited for large configuration spaces and spaces that include invariant constraints.


\section{Neural Motion Planning} 

\begin{figure}[tb]
    \centering
    \includegraphics[width=0.32\textwidth]{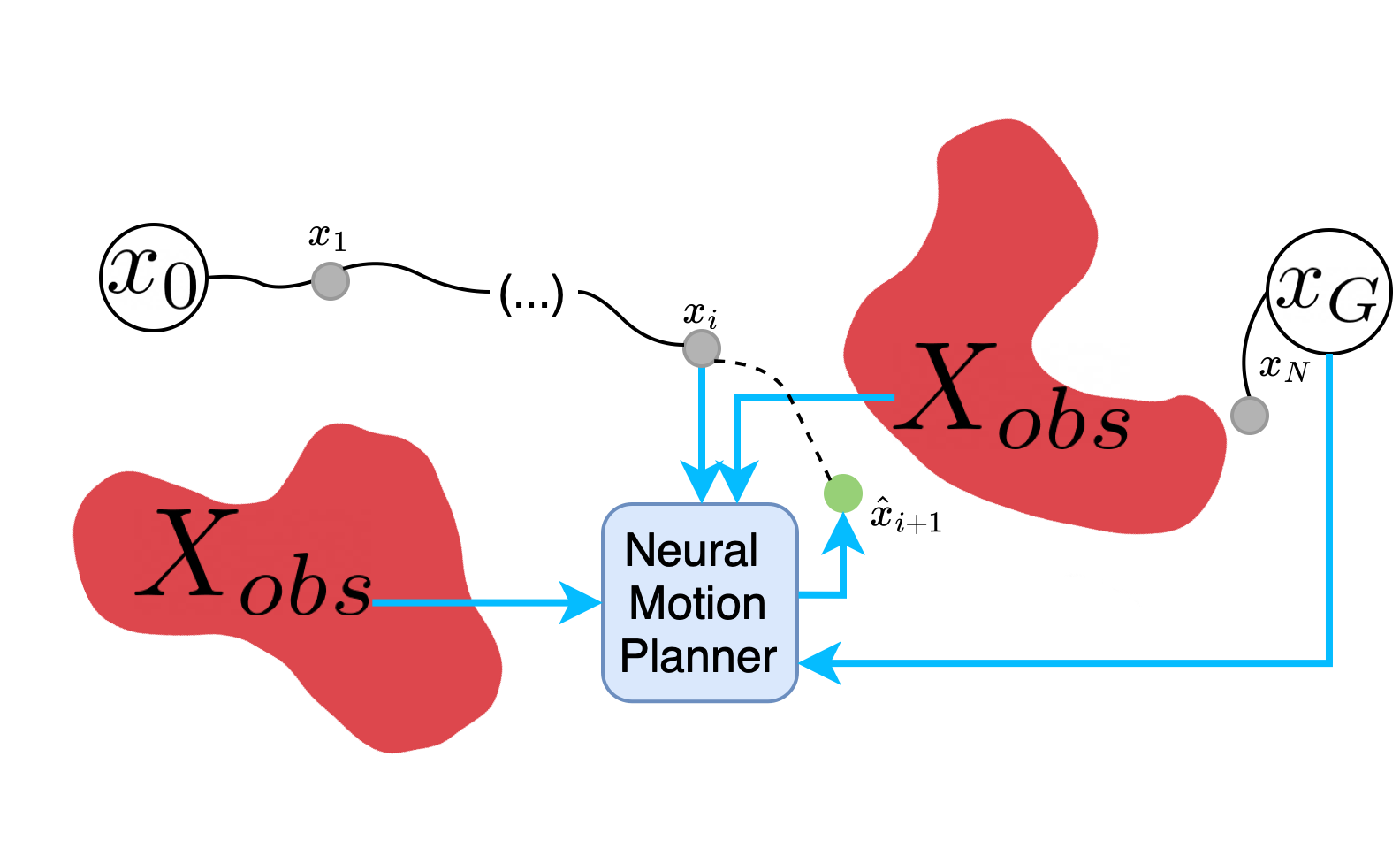}
    \includegraphics[width=0.32\textwidth]{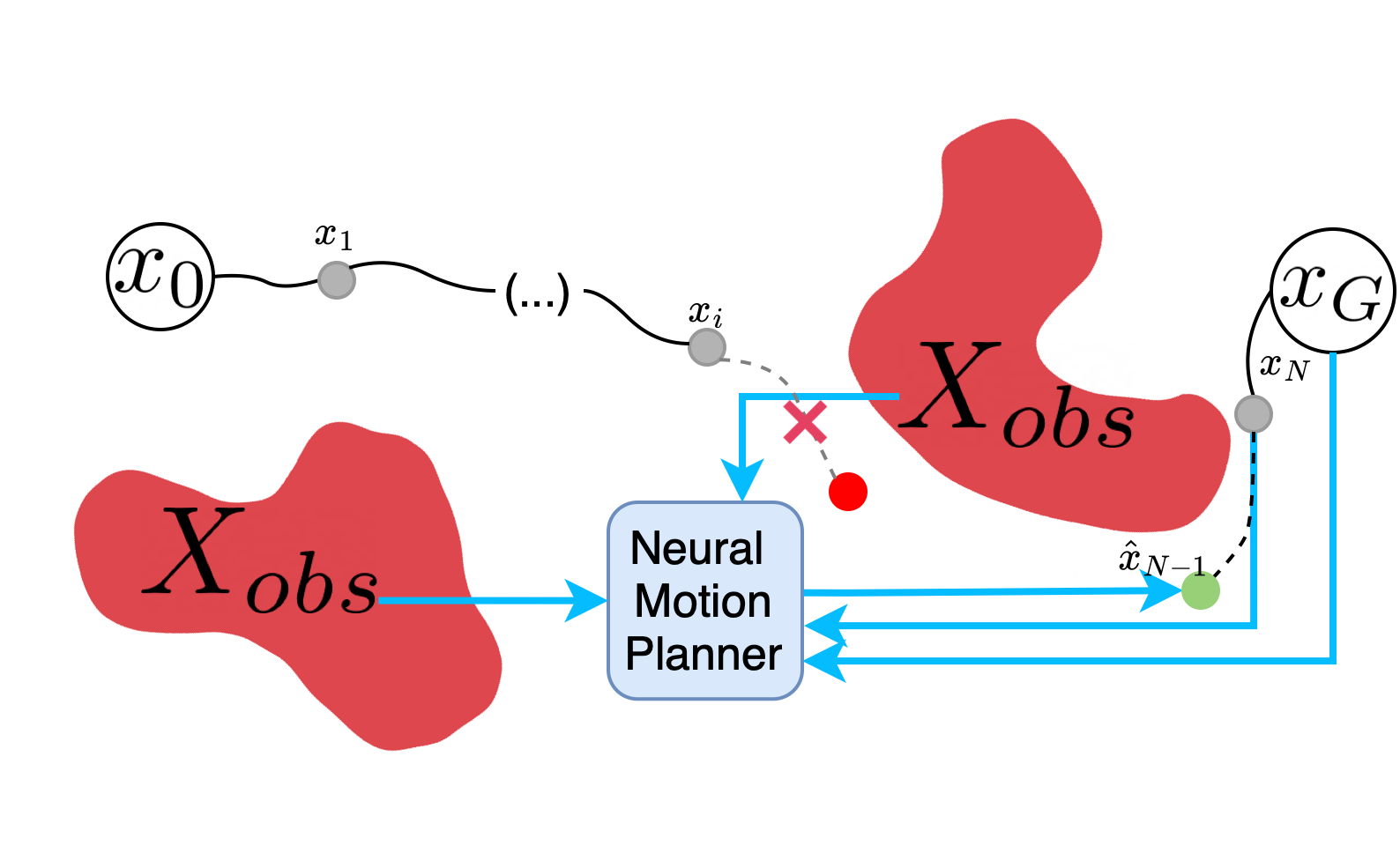}
    \includegraphics[width=0.32\textwidth]{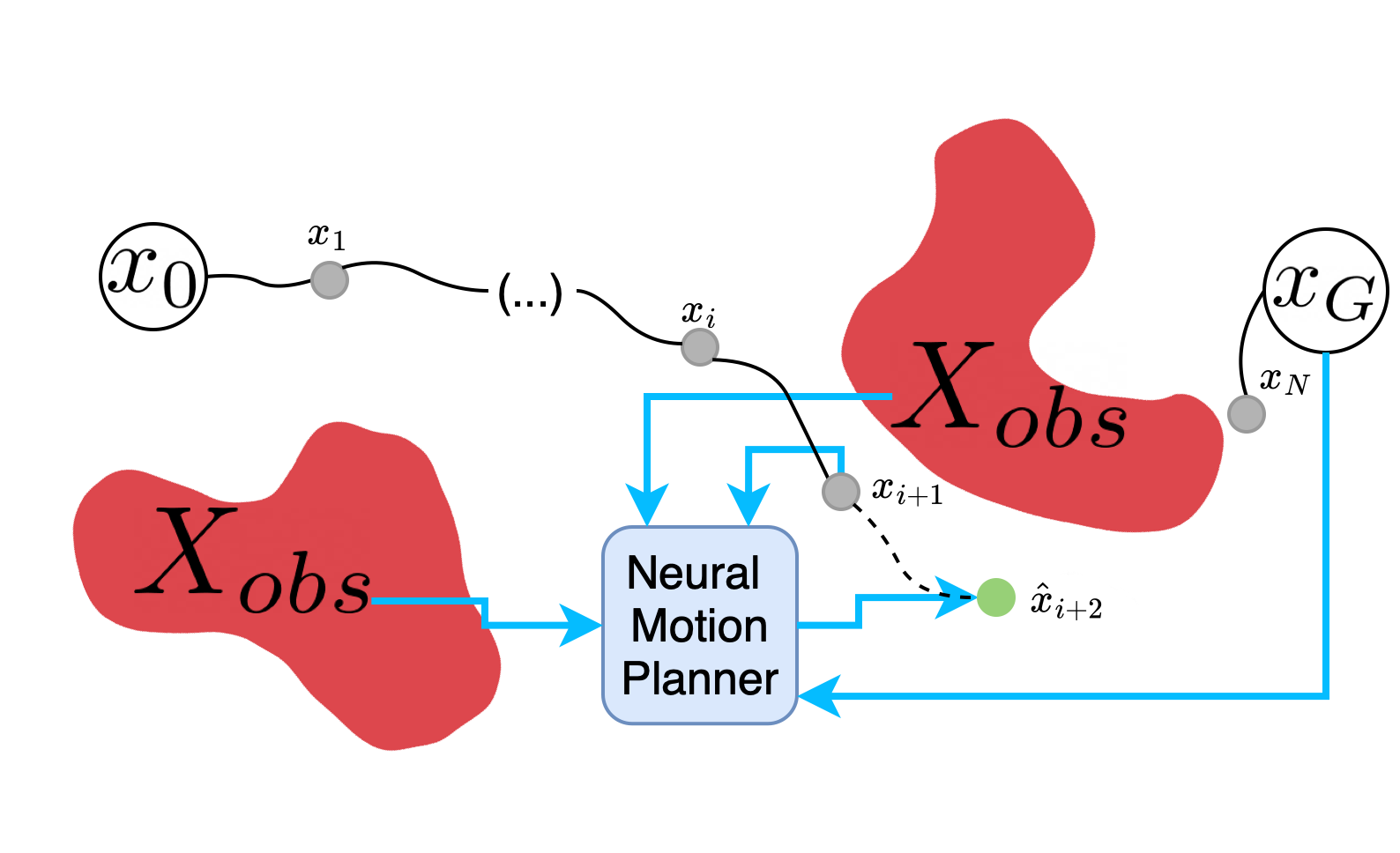}
    \caption{{\bf Architecture of a neural motion planner (NMP) \citep{qureshi2019motion}.}  The NMP takes as input the current state $x_i$, the goal state $x_G$ and a representation of the environment such as a point cloud. (Left)  The NMP outputs $\hat{x}_{i+1}$, the predicted next state of the robot at the next timestep, and attempts to connect to it. (Middle) If the connection fails, a path from the goal is attempted. (Right) If the connection succeeds, the predicted state is added to the solution path. A hybrid approach uses the NMP alongside a traditional \sbmp \ like RRT*.}
    \label{fig:neural_motion_planner}
\end{figure}

This class of methods uses a dataset of solved motion planning problems to learn a neural network model that improves the efficiency of motion planning on different, yet similar problems (Figure~\ref{fig:neural_motion_planner}). \emph{Motion Planning Networks (MPNet)} \citep{qureshi2019motion} is a deep neural network architecture which learns to approximate the computation of a sampling-based motion planner. MPNets are comprised of an  encoder  network  and  a planning  network.  The encoder network takes as input a point-cloud  representation of a workspace and learns to encode this point-cloud into a latent space.  The planning network takes as input the latent space, the robot's configuration at the current time-step and a goal configuration, and it is trained to predict the robot's configuration at the next time-step.  The planning network is used along with a bi-directional iterative search algorithm 
to generate trajectories that are feasible given the problem's constraints.  

\textit{Deep Deterministic Policy Gradient for Motion Planning} (DDPG-MP) \citep{JurgensonT19} investigates an alternative approach to training a motion planner approximated as a neural network (referred to as a \textit{neural motion planner}). The approach is based on reinforcement learning rather than supervised learning, and achieves higher accuracy by actively exploring the problem domain. 

An integral component of neural motion planners is the learned representation of the problem's obstacles from sensory input. Recently, the \textit{PointNet} architecture \citep{qi2017pointnet} has been used to learn an efficient and effective representation of point clouds for neural motion planning \citep{strudel2020learning}. This representation is then concatenated with the goal configuration and passed to a neural network which outputs actions. This network uses the \textit{Soft Actor-Critic} (SAC) reinforcement learning algorithm to learn the control policy online.

\textit{Dynamic MPNet} \citep{dynamic-mpnet} extends the MPNet framework to non-holonomic robots. The network is trained on example solution trajectories produced by an RRT* planner to predict the next state that the robot must steer to. The network receives as input the goal state and a costmap of the local area surrounding the robot's current state. The neural planner is called iteratively at every state until it produces a feasible solution trajectory. If no such trajectory is found, the method falls back to classical motion planning.

\textit{MPC-MPNet} \citep{li2021mpc} extends the MPNet framework to deal with the challenges of kinodynamic planning. The network is trained on example solution trajectories produced by the AO SST algorithm \citep{li2014asymptotically} to predict a batch of candidate next states (waypoints) given a batch of environment encodings as well as current and goal states. Parallelized Model Predictive Control (MPC) is used to steer the robot from its current state to the predicted waypoints. The procedure is repeated either in a greedy or tree-based search framework to obtain kinodynamically feasible solution trajectories in new planning environments.

\textit{Constrained MPNets} (CoMPNet) \citep{9501956} and its extension, CoMPNet X \citep{9143433} extend the MPNet framework to the problem of motion planning in the presence of task constraints. CoMPNet takes as input a task and observation encoding and outputs an intermediate configuration for path planning. The demonstrated trajectories are collected using a bi-directional RRT planner. CoMPNetX introduces neural-gradient-based projections to generate informed, implicit manifold configurations that can speed up any \sbmp.

\section{Learning a lower dimensional planning space}

\begin{figure}[tb]
    \centering
    \includegraphics[width=.99\textwidth]{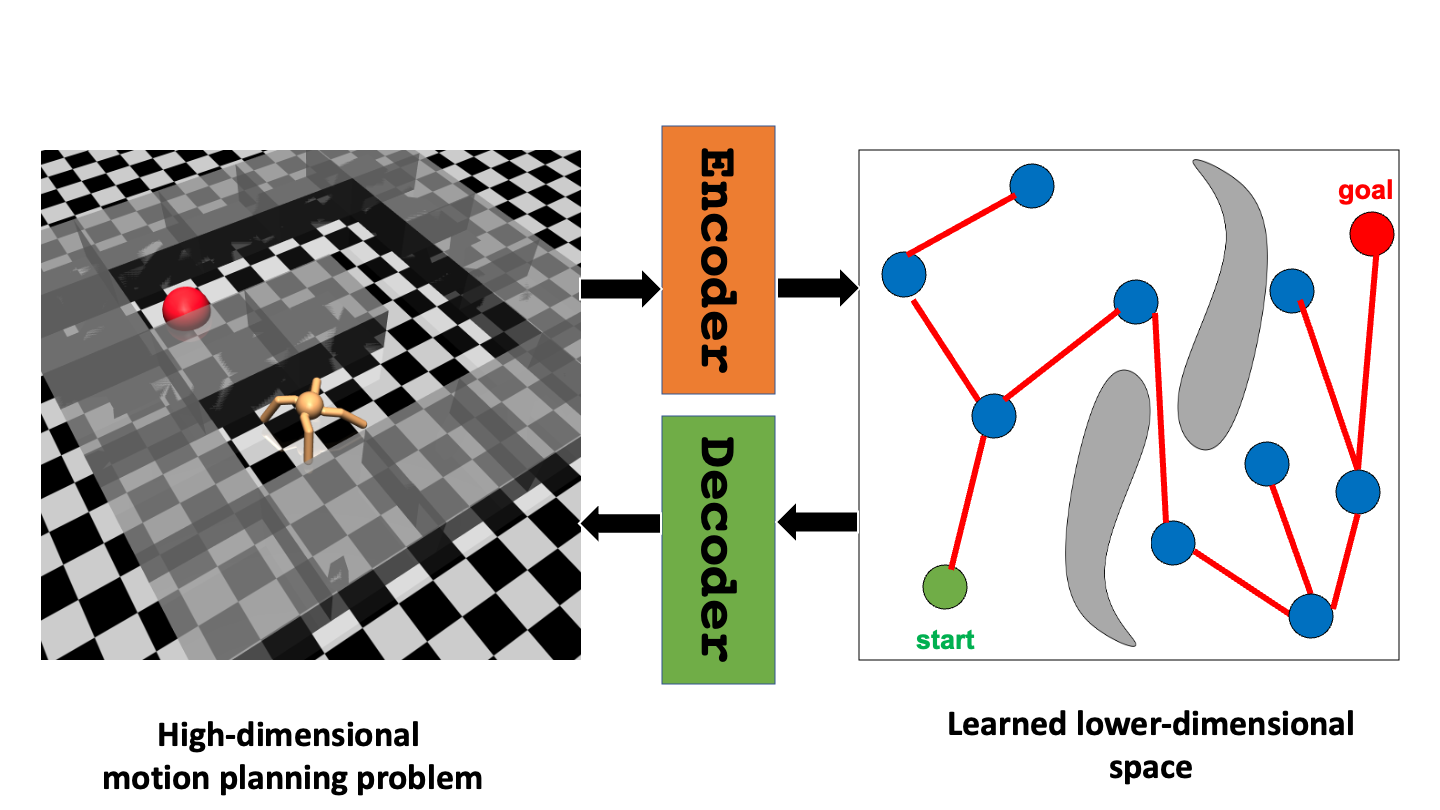}
    \caption{{\bf Example of using learned lower dimensional space. } A simulated quadruped robot (left) with a 111-dimensional state space and 8-dimensional control space must navigate the maze to reach the goal (red circle). It may be practically infeasible to apply a \sbmp \ directly to this problem, so a learned \textit{encoder} model transforms the problem into a lower-dimensional space (right) where a \sbmp \ can be applied to find a solution. Once a solution has been found, it can be applied directly to the high-dimensional problem via a learned \textit{decoder}.}
    \label{fig:lower_dimensional_space}
\end{figure}

Under this relatively new framework, a learned model maps a high-dimensional motion planning problem into a lower dimensional state space (Figure \ref{fig:lower_dimensional_space}), and uses SBMP techniques to solve the lower dimensional problem. \textit{Learned Latent RRT} (L2RRT) \citep{ichter2019robot} learns this mapping via an autoencoder. It constructs a tree directly in this lower dimensional space by propagating randomly sampled controls in the lower dimensional space using a learned latent dynamics model, and checking for collisions in the lower dimensional space using a learned collision checking neural network. The autoencoder and latent dynamics model are trained using trajectories obtained by iteratively propagating random samples from the control space for a fixed number of time steps. The collision checking network is trained using randomly samples from low dimensional states (obtained via the trained encoder) that are in $\mathbb{C}_{free}$ and $\mathbb{C}_{obs}.$ The collision checking network outputs the probability of a latent state $z_t$ being in collision or not, and L2RRT makes use of a pre-defined confidence threshold to decide whether to trust the learned collision checker or not.

The \textit{Neural EXploration-EXploitation Tree} (NEXT) \citep{chen2019learning} is a meta neural motion planner that learns generalizable problem structures from previous planning experiences. An attention-based neural network architecture is trained to extract a planning representation from demonstrated paths on multiple random problems. The learned tree expansion procedure can balance exploration and exploitation by using an Upper Confidence Bound (UCB) style algorithm.

\textit{Broadly-Exploring, Local-policy Trees} (BELT) \citep{ichter2021broadlyexploring} is an RRT-like algorithm for long-horizon task planning that plans over an abstract task space, which is learned from demonstrations. The exploration of the RRT is guided by a task-conditioned, learned policy capable of performing general short-horizon tasks. Once a solution is found by using search, the local task-conditioned policy is capable of executing the trajectory on a real robotic platform.

\begin{mdframed}[hidealllines=true,backgroundcolor=blue!20,frametitle=Discussion on Learning-based Pipelines]
There is a growing body of work in collecting prior motion planning experience, and using learned models to imitate the motion planner's performance on tasks from a similar domain. These pipelines may also project a higher dimensional problem into a reduced space, where it may be easier to find a feasible path. These pipelines can result in practical efficiency by reducing expensive SBMP operations to a few inference calls to a learned model. Nevertheless, there are considerations about the size of the datasets required to train these models, their generalization capabilities, as well as theoretical guarantees on path quality, that remain open directions for future research. 
\end{mdframed}

\chapter{SBMP with Learned Models}
\label{sec:planning_under_uncertainties}



This section looks into how machine learning has been used to handle, and plan under, uncertainty. Noisy sensors, actuators and external agents are potential sources of uncertainty. Sampling-based planners have been applied to planning under uncertainty through belief space planning. As with kinodynamic planning, many belief-space planning problems do not have access to a steering function, and thus planning must be done using only forward propagation. Uncertainties can also arise from the unavailability to measure parameters like wheel-ground friction, air friction, spring constants etc. 

\begin{mdframed}[hidealllines=true,backgroundcolor=red!20,frametitle=Categories on SBMP with Learned Models]
\begin{myitem}
    \item \textbf{Modeling robot uncertainty} \citep{osti_10145334, 9196564,McConachie_2020, malone2012implementation,doi:10.1177/02783649211004615,curtis2022long,bahnemann2017sampling,burri2018framework,omnirobot}
    \item \textbf{Modeling uncertainty in obstacle dynamics} \citep{Aoude2013ProbabilisticallySM,88149,fulgenzi2008probabilistic,fulgenzi2010risk,zhang2020novel}
\end{myitem}
\end{mdframed}

\section{Modeling Robot Uncertainty}
Learned models have been used for systems where precisely modeling the motion of the objects being manipulated is infeasible (for instance, manipulation with underactuated hands) \citep{osti_10145334}.  This method learns a \textit{stochastic} model of the system's dynamics from a dataset of trajectories with randomly sampled actions. Two choices of stochastic models - a Bayesian neural network and a Gaussian Process (GP) are explored. The trained dynamics model is used with a sampling-based planner that represents beliefs as a particle cloud over the state space. The model forward propagates these particle clouds in order to construct a planning tree. Critic models have also been applied in this domain to estimate the error in the learned transition model \citep{9196564}. The output of the critic model is used as a cost function by an AO planner to direct planning towards regions where there is less uncertainty in the learned dynamics. 

Learned models have also been applied to problems where the underlying system may be difficult and/or time consuming to model correctly, and thus the planner acts over a \textit{reduced} state space \citep{McConachie_2020}. Training data for this model is collected by generating plans in the reduced state space, and then applying these plans to the true system. This data is used to train a classifier that labels plans generated in the reduced state space as reliable or unreliable.  The learned model is incorporated into an RRT-based planner in order to bias sampling away from transitions that are labeled as unreliable.

The \textit{Brain-Emulating Cognition and Control Architecture} (BECCA) \citep{malone2012implementation} uses online reinforcement learning to exploit task structure as well as to address environmental noise and hardware imperfection inherent to manipulation tasks. At every time step, BECCA makes an observation in the world, extracts features from the sensory input, performs an action in response to the input, and receives a reward. The feature extractor identifies patterns and correlations in the input vector in an unsupervised manner. The action selector learns, via online reinforcement learning, a mapping from features to actions that lead to the highest recorded reward. The architecture is trained on a simulated system until it obtains satisfactory performance, and is allowed to update its model after being deployed. It is mapped to a PRM by reducing the agent's state space to roadmap nodes, and limiting the agent's actions to edges between the nodes.

Conditions for pouring or scooping skills are learned using GPs to enable risk-aware predictions \citep{doi:10.1177/02783649211004615}. Alongside active learning, this approach reduces the model's uncertainty. The learned skills are then used by a \textit{RRT-Connect} \citep{kuffner2000rrt} to produce geometric paths for the gripper. A related approach \citep{curtis2022long}, explores ways to implement perceptual learned modules for shape estimation and grasp generation that are then used by a SBMP.


Hybrid data-driven approaches have been used to learn models for robots with imprecise or uncertain dynamics. These techniques combine physics models of a robot with machine learning methods to self-tune the physics models, in order to accommodate robot design imprecision. A crucial component is generating valuable data in order to reduce the tuning time. For this purpose, a modified RRBT planner is used to search for \textit{informative trajectories}, those which maximize the information gain of the specified unknown parameters \citep{bahnemann2017sampling}. This method is used for calibrating a micro aerial vehicle in a constrained environment with minimal user interaction, and is later applied in \cite{burri2018framework} to produce trajectories for accurate parameter estimation based on maximum likelihood. Another system identification approach is to, given the dynamical model of the system, learn the unknown friction coefficients. The learned model is used by a SBMP to generate more reliable trajectories \citep{omnirobot}. 

\section{Modeling Uncertainty in Obstacle Dynamics}
\textit{RR-GP} \citep{Aoude2013ProbabilisticallySM} applies learned Gaussian Processes (GPs) for probabilistic safe motion planning with dynamic obstacles and uncertain motion patterns. RR-GP combines GPs with RRT-Reach to build a learned motion pattern model that can be used to predict obstacle trajectories.  These predictions are conditioned on feasible paths which are identified using reachability analysis.  RR-GP uses a chance-constrained RRT to identify probabilistically feasible paths. 

Hidden Markov Models (HMMs) have also been used to model the possible motion of dynamic obstacles \citep{88149}.  Obstacles are modeled as a stochastic process within the HMM, which is queried to obtain obstacle motions and potential variations of the motion states.  Planning is performed using a trajectory-guided path-planning algorithm which samples and evaluates a set of candidate trajectories.  

Learning has been used to predict motions of dynamic obstacles, such as pedestrians or vehicles, in order to reduce the risk of collision \citep{fulgenzi2008probabilistic,fulgenzi2010risk}. Motions of obstacles are modeled using GPs, which are learned by observing an environment to capture common behaviors and patterns.  This model provides predictions, which are continuous over space and time, and when applied to planning gives better performance than kinematic-based predictions.

In a similar approach, \cite{zhang2020novel} uses SBMP in the context of autonomous driving. This work proposes labeling and training models to more accurately predict a vehicle’s intention, in order to sample states leading to a high-quality collision-free trajectory. Prediction of surrounding vehicles and imitation learning are used to generate collision-free samples near the human-driving trajectory that are used by the planning algorithm.

\begin{mdframed}[hidealllines=true,backgroundcolor=blue!20,frametitle=Discussion on Planning with Learned Models]
        Machine learning models, such as Bayesian Neural Networks and Gaussian Processes (GPs), allow to reason about uncertainty in a learned model. When SBMP techniques are applied to problems involving robots in the real world, they have to often deal with physical processes that cannot be modeled perfectly. A promising area of research is to use these stochastic machine learning models to improve the efficiency and accuracy of sampling-based planners in a way that allows for the underlying uncertainty of the model. 
\end{mdframed}



\chapter{Discussion}

This monograph studied how machine learning has been used to improve the performance of \sbmps. This can be achieved by using machine learning models to approximate the primitive operations of a sampling-based planner or to select between different choices of primitive operations. It also discussed a relatively new area of research, where machine learning models have been proposed that approximate the operation of an optimal sampling-based planner. Finally, it briefly covered some of the literature where sampling-based planning can be performed over a learned robot and/or environment model.

Many \sbmps  \ exhibit desirable theoretical properties, such as probabilistic (or resolution) completeness and asymptotic optimality, but may be slow to converge to high quality solutions in practice. One promising direction to improve the quality of solutions is to work within the \sbmp \ framework and provide desirable guarantees, while improving the practical convergence rate by using efficient operations based on machine learning models. This monograph concludes by briefly evaluating the successes (and failures) of the different proposed integration techniques, in the context of the practical challenges described in Section~\ref{sec:intro}.

\section{Computational Efficiency}

The first form of computational efficiency that bears consideration is that of \emph{offline} vs \emph{online} computation. For a machine learning model to capture relevant features of a motion planning problem instance, either in a single workspace or across workspaces, it must have access to a high-quality dataset that does not lead to a significant statistical bias. 

Some works discussed in this survey trade an environment-specific, offline cost, to obtain better performance on new planning problems in the same environment. Alternatives trade a higher offline cost to obtain better performance on new environments (that are similar to the ones seen during training). Deciding which of these trade-offs to use depends on the application where the planner is being deployed, such as whether the environment the robot will be deployed is known or not. 

For some problems, such as learning a sampler via supervised learning or learning a collision-checker inside a workspace, it is straightforward to annotate high-quality datasets via significant offline computation. The trained model rewards this offline effort by speeding up the runtime of the planner on a new problem instance. Nevertheless, this survey discussed applications, such as approximating a distance function or sampling strategies for manipulation planning, where obtaining such a dataset is computationally more challenging, which motivates additional research effort in these areas.

A relatively unexplored area of research is the idea of an \sbmp \ that learns from previous planning experience on similar, yet different, problems, and manages to transfer the experience across environments. Such a planner would spend more computational time at the beginning of its life cycle but would eventually gather enough planning experience to find high-quality solutions to new problems quickly. 

A form of computational efficiency that must also be taken into account is online inference time, e.g., when the learned module inside an \sbmp\ uses a large machine learning model, such as a deep neural network. Such an integration may lead to an improvement over an alternative planner in terms of the number of iterations to find a solution. For most practical purposes, however, the metric of interest is \textit{overall planning time}, where the integration of an \sbmp\ with a learned model may under-perform due to inference time. Specialized hardware, such as GPUs, can perform model inference for large machine learning architectures to obtain significant gains. This may not be practical, however, in all applications.

\section{Path Quality}

For graph-based \sbmps, such as the PRM, the most significant determiners of path quality are the sampling method and the nearest neighbor selection, which is performed according to a distance metric. Learned samplers discussed in Section~\ref{sec:sampling} produce samples along the shortest path and show the most promise for improving path quality. Learning accurate distance metrics, and integrating them efficiently into nearest neighbor search subroutines, remains an open area of research.


For tree-based \sbmps, such as the RRT, in addition to node selection, node expansion majorly impacts path quality. Expansion typically involves randomly sampling a state, and then querying a local planner to make progress towards the selected state. This opens the door for learned models to both identify good states to sample (i.e., are likely to lie along a shortest path), and good local plans that bias the final path towards lower costs.


Irrespective of the underlying \sbmp \ being used, it would be interesting to explore the application of machine learning to perform post-processing on paths computed by an \sbmp. Just like machine learning models have been used to identify the order in which edges must be collision-checked, a model could be trained to identify the segments of a tree or a graph that are good candidates for smoothing, or to apply the smoothing directly. Potentially, this can be achieved in conjunction with an optimization process. 



\section{Ease of Use}
One consideration during the deployment of \sbmps \ is that they involve a set of parameters that need to be tuned. Furthermore, the sampling strategy, distance metric, collision detection method, and connection strategies, are all exchangeable components. The selection of these primitives as well as parameter values is non-trivial, and often requires expert knowledge of motion planning for the specific application. Inappropriate selection can handicap the performance of the planner.


Adaptive methods that automatically select combinations of primitive operations and parameter settings offer a solution, which allows non-expert users to make use of \sbmps\ effectively.  Previous work includes adaptive methods for selecting samplers, connection strategies and parameter settings. It would be interesting and useful to combine these methods to create \emph{universal} planning frameworks that use a learned model to adaptively select combinations of primitive operations and parameter settings.  


\section{Limitations of Existing Methods}


Successful motion plans are a rich source of data about multiple facets of the environment or that particular planning instance, including distance between robot configurations, and the validity of different parts of the workspace. The majority of the approaches discussed in Section~\ref{sec:learning-primitives}, however, deal with only a single primitive replaced with a learned module. On the other side of the spectrum, the neural motion planners in Section~\ref{sec:integrated} replace the entire \sbmp \ with learned models. There has not been enough exploration in the literature that focuses on the interaction of multiple learned components within an \sbmp\ and for which purpose they offer the most benefits. 


Different machine learning models expose different types of \emph{inductive bias}, i.e., the ability of the learning algorithm to predict outputs given inputs that it has not encountered during the training process. Relatively new architectures, such as Graph Neural Networks (GNNs) \citep{Bruce_2014} and Transformers \citep{vaswani2017attention}, have been used in a wide variety of applications in computer vision and natural language processing. These models may enable similar advancements in improving the performance of \sbmps.

Poorly designed and inadequately trained models will significantly handicap planning performance.  Thus, it becomes increasingly important to develop motion planning algorithms that are robust to inaccurate models. A learned model can also potentially adversarially impact the probabilistic guarantees of \sbmps.  Many works discussed in this survey do not discuss the impact of machine learning on the properties (probabilistic completeness or asymptotic optimality) of the \sbmp\ framework. For example, if a learned sampler only generates samples over a subset of $\cspace$, then any \sbmp\ that uses this sampler will not be complete.  It would be interesting to develop learning-based methods that provide theoretical guarantees.

\section{Potential Future Work}

In most robotic applications, there is the consideration of integrating the planning algorithm with the robot's perception. Although machine learning promises end-to-end pipelines that output low-level control policies for a given task, given raw sensory input, these are often data-hungry solutions, and they do not easily generalize across domains and tasks. Nevertheless, there are a lot of opportunities in using machine learning models to learn high-level, task-specific, and interpretable representations of a robot's sensory input, which a downstream \sbmp \ can use.

As there are many avenues for integrating machine learning tools with \sbmps, it is also useful to define clear benchmarks that can evaluate the performance, and highlight the strengths and weaknesses of a new motion planner across a variety of problems. There has been preliminary work in this area \citep{chamzas2021motionbenchmaker}.

Many successful algorithmic solutions that integrate machine learning tools with \sbmps \ exploit the inherent structure present in specific applications, and design a machine learning model that can solve a problem component. Examples in this survey include the design of a learned local planner for systems with significant dynamics, and learning stochastic dynamics models for planning under uncertainty. These solutions are often handcrafted to specific applications, but they offer promising avenues for motion planning researchers with domain expertise to investigate.

\section{Alternative Planning Frameworks}

Although outside the scope of this survey, there are other approaches to the motion planning problem beyond \sbmps. These include search-based approaches, which search for paths from start to goal configurations by combining a series of \textit{motion primitives} (short, kinematically feasible motions). They also include trajectory optimization methods \citep{ratliff2009chomp}, that assign a cost to each robot trajectory and incorporate constraints, such as path smoothness and collisions. They can use a broad class of optimization techniques, such as sequential convex optimization, to obtain a good solution trajectory. There are also efforts that apply machine learning to these alternative planning methodologies \citep{Vemula-RSS-20,mdydb-2018}.

\backmatter  

\printbibliography

\end{document}